\documentclass[5p,times,twocolumn]{elsarticle}

\usepackage{geometry}
\usepackage[T1]{fontenc}
\usepackage[utf8]{inputenc}
\usepackage{amsmath,amssymb,amsfonts}
\usepackage{algorithmic}
\usepackage{graphicx}
\usepackage{textcomp}
\usepackage{xcolor}
\usepackage{booktabs}
\usepackage{placeins}
\usepackage{subcaption}
\usepackage{hyperref}
\usepackage[figuresright]{rotating}


\journal{Knowledge-Based Systems}

\begin{document}

\begin{frontmatter}

\title{Kolmogorov-Arnold Networks vs. Multi-Layer Perceptrons: A Rigorous Statistical Benchmark on Clinical Tabular Data}

\author[aff1,aff2]{Rafa\l{} Wodzi\'nski\corref{cor1}}
\ead{rafal.wodzinski@student.pk.edu.pl}

\author[aff1,aff3]{Pawe\l{} Szczerbiak\corref{cor1}}
\ead{p.szczerbiak@sanoscience.org}

\cortext[cor1]{Corresponding author}

\affiliation[aff1]{organization={Sano Centre for Computational Medicine},
            city={Krak\'ow},
            country={Poland}}

\affiliation[aff2]{organization={Department of Computer Science, Faculty of Computer Science and Mathematics, Cracow University of Technology},
            city={Krak\'ow},
            country={Poland}}

\affiliation[aff3]{organization={Faculty of Computer Science, AGH University of Krakow},
            city={Krak\'ow},
            country={Poland}}

\begin{abstract}
Kolmogorov-Arnold Networks (KANs) have recently been proposed as a promising alternative to Multi-Layer Perceptrons (MLPs). However, empirical evaluations on structured medical datasets remain scarce, and commonly reported metrics like AUROC can yield overly optimistic performance estimates on imbalanced clinical cohorts. In this study, we present a standardized benchmark comparing eleven architectures---nine KAN variants (FastKAN, WavKAN, ReLUKAN, ChebyKAN, TaylorKAN, GramKAN, JacobiKAN, LegendreKAN, and TabKAN), a tuned StandardMLP, and a Random Forest baseline---across seven heterogeneous medical tabular datasets spanning oncology, cardiology, nephrology, neurology, endocrinology, and obstetrics. Our experimental protocol employs a $3\times5$ Repeated Stratified $K$-Fold cross-validation scheme with nested Bayesian hyperparameter optimization and evaluates discrimination via Area Under the Precision-Recall Curve (AUPRC), calibration through the Brier Score, and computational efficiency. Formal equivalence testing is conducted using the Bayesian Correlated $t$-Test with a Region of Practical Equivalence (ROPE) at $\pm 0.01$. The results indicate that on datasets with severe class imbalance, dedicated KAN architectures (particularly TabKAN and ReLUKAN) can improve minority-class detection and achieve higher AUPRC than standard MLPs. Nevertheless, across pooled datasets, KAN variants demonstrate comparable overall predictive discrimination rather than a systematic advantage, while requiring $1.20\times$ to $1.48\times$ longer inference latency per sample and up to $12.5\times$ more trainable parameters. Consequently, StandardMLP remains a strong baseline on the efficiency Pareto frontier, while KAN variants provide a viable alternative when precision on rare positive cases is the primary clinical objective. The code for reproducing the results is available at \url{https://github.com/rafalwodzinski/kan_mlp_comparison}.
\end{abstract}


\end{frontmatter}

% ===========================================================
% SECTION 1: INTRODUCTION
% ===========================================================

\section{Introduction}
\label{sec:introduction}

Structured tabular data represents the primary modality in clinical medicine. Electronic health records, laboratory test panels, patient registries, and continuous physiological streams support diagnostic decisions, treatment selection, and patient triage across health systems. Machine learning models deployed in clinical decision support must satisfy three practical requirements simultaneously: high discriminative performance, reliable probability calibration, and low inference latency suitable for interactive workflows \cite{rajpurkar2022ai}. These requirements often conflict. For example, uncalibrated risk predictions can cause misleading probability estimates, potentially compromising patient safety in critical care environments \cite{van2019calibration}. Furthermore, real-world clinical datasets almost invariably exhibit severe class imbalance, where positive disease cases represent only a small minority of the cohort. In such settings, standard aggregate metrics such as the Area Under the Receiver Operating Characteristic Curve (AUROC) can provide an overly optimistic evaluation. Because AUROC incorporates the large pool of true negatives into the false positive rate, it can mask high false discovery rates. Consequently, the Area Under the Precision-Recall Curve (AUPRC) has emerged as an essential evaluation metric, as it concentrates directly on the retrieval and precision of the minority class \cite{saito2015precision}.

Empirical studies consistently indicate that tree-based ensembles---primarily XGBoost \cite{chen2016xgboost} and Random Forests \cite{breiman2001random}---remain the dominant choice for tabular machine learning benchmarks \cite{shwartz2021tabular,grinsztajn2022why}. Several factors explain this phenomenon: tabular features typically possess independent semantic meanings, missing values are common, and target decision surfaces are often sharp and non-smooth. Traditional deep neural networks, by contrast, rely on inductive biases such as rotational invariance and spatial correlation, which can impair performance on heterogeneous tabular inputs. Nonetheless, Multi-Layer Perceptrons (MLPs) \cite{hornik1989multilayer} remain the standard neural baseline for tabular data due to their simple design, fast inference, and direct compatibility with standard GPU and accelerated hardware kernels.

Recently, Liu et al.\ \cite{liu2024kan} introduced Kolmogorov-Arnold Networks (KANs) as an alternative neural framework based on the Kolmogorov-Arnold representation theorem \cite{kolmogorov1957representation,arnold1957functions}. Instead of placing fixed activation functions at nodes, KANs position learnable univariate activation functions on the network edges. In the original implementation, these functions were parameterized via B-splines. While KANs demonstrated promising accuracy on symbolic regression and mathematical toy problems, they exhibited a substantial computational drawback: the original authors reported roughly a tenfold increase in training time relative to MLPs of comparable width \cite{liu2024kan}, driven by the sequential evaluation of spline bases. To mitigate this computational cost, several alternative KAN variants have been proposed: FastKAN approximates splines via Gaussian Radial Basis Functions \cite{li2024fastkan}; Wav-KAN adopts wavelet transforms \cite{bozorgasl2024wavkan}; ReLU-KAN reconstructs spline-like activation curves using native linear and ReLU operations \cite{qiu2024relukan}; ChebyKAN applies Chebyshev polynomial expansions \cite{ss2024chebykan}; TaylorKAN combines Taylor series expansions with dimensionality reduction \cite{chen2024taylorkan}; and related studies explore orthogonal functional bases such as Jacobi \cite{aghaei2024fkan, spacelearner2024jacobikan, XIONG2026109414}, Legendre \cite{HU2026109388}, and Gram \cite{khochawongwat2024gramkan} polynomials. More recently, TabKAN \cite{eslamian2025tabkan} was designed specifically for tabular domains by combining KAN layers with feature gating mechanisms.

Despite active methodological interest, existing evaluations of KAN architectures feature notable limitations. Most published studies evaluate models primarily on synthetic equations, vision benchmarks such as MNIST, or small non-medical datasets. Few works examine KANs on heterogeneous medical tabular data with class imbalance, and existing reports rarely evaluate threshold-free metrics like AUPRC or quantify predictive calibration. Furthermore, published comparisons frequently report raw point estimates from single cross-validation runs without formal statistical equivalence testing. As a result, whether KANs offer a meaningful advantage over tuned MLPs on clinical tabular tasks remains an open empirical question.

To address this gap, we present a systematic empirical benchmark comparing eleven model architectures (nine KAN variants, a StandardMLP baseline, and a Random Forest ensemble) across seven medical tabular datasets spanning diverse clinical domains. Our experimental design applies a 3$\times$5 Repeated Stratified $K$-Fold cross-validation protocol with nested Bayesian hyperparameter optimization (via Optuna \cite{akiba2019optuna}) and strict fold-isolated preprocessing to prevent information leakage. We assess models across discrimination (AUPRC, AUROC, Matthews Correlation Coefficient, F1-Score), calibration (Brier Score), and computational cost (per-sample inference latency, training time, parameter count). In addition, we apply the Bayesian Correlated $t$-Test with Region of Practical Equivalence (ROPE) \cite{benavoli2017time,corani2015bayesian} to test whether observed performance differences represent practical equivalence or meaningful improvements.

The results of our benchmark provide two primary insights:
\begin{enumerate}
    \item On datasets characterized by severe class imbalance (such as cervical cancer screening and cardiotocography), specific KAN variants---notably TabKAN and ReLUKAN---achieve competitive or superior AUPRC and minority recall relative to the StandardMLP baseline.
    \item Across the broader pooled benchmark, the discriminative performance and calibration of KAN variants and the StandardMLP are largely equivalent within a $\pm 0.01$ ROPE interval. However, all evaluated KAN variants incur a noticeable computational penalty: per-sample inference latency is $1.20\times$ to $1.48\times$ higher, and parameter counts are $2.8\times$ to $12.5\times$ larger than those of the StandardMLP.
\end{enumerate}
These findings suggest that while standard MLPs remain the more computationally efficient choice for general clinical applications on the Pareto frontier, tailored KAN architectures warrant consideration when maximizing detection of rare clinical endpoints is of primary importance.

% ===========================================================
% SECTION 2: RELATED WORK
% ===========================================================

\section{Related Work}
\label{sec:related_work}

\subsection{Kolmogorov-Arnold Networks and Their Variants}
\label{subsec:rw_kan}

Kolmogorov-Arnold Networks were introduced as a mathematically principled alternative to the Multi-Layer Perceptron, grounded in the Kolmogorov-Arnold Representation Theorem. In place of the fixed nodal activation functions and learned scalar weights of the standard MLP, KANs place learnable univariate functions---parametrised as B-splines---on the edges of the network graph, yielding a representation of the form $f(\mathbf{x}) = \sum_q \phi_q \left( \sum_p \phi_{q,p}(x_p) \right)$. The original work demonstrated reduced parameter counts at comparable approximation error on symbolic regression and physics simulation benchmarks, and proposed interpretable Pareto frontier analyses suggesting favourable scaling relative to MLPs. However, the original authors themselves explicitly conceded that KANs are ``currently 10$\times$ slower to train than MLPs for a given number of parameters'' \cite{liu2024kan}, attributing this bottleneck to the sequential nature of B-spline computations, which cannot exploit GPU tensor parallelism as efficiently as dense matrix multiplications. No evaluation on real-world structured tabular data was conducted.

The computational bottleneck of the original architecture immediately motivated a series of alternative formulations. \textsc{FastKAN} \cite{li2024fastkan} demonstrated that third-order B-splines can be effectively approximated using Gaussian Radial Basis Functions, yielding a $3.33\times$ forward-pass speedup over \texttt{efficient\_kan} (though the combined forward-backward speedup shrank to $1.25\times$), while noting in a footnote that MNIST is not an optimal setting for evaluating KAN capabilities. \textsc{Wav-KAN} \cite{bozorgasl2024wavkan} proposed wavelet transforms as edge activations for noise robustness, though its parameter complexity across an $L$-layer network is $\mathcal{O}(3N^2L)$ (accounting for scale, translation, and weight parameters) compared to $\mathcal{O}(N^2L)$ for a standard MLP of equivalent width $N$. \textsc{ReLU-KAN} \cite{qiu2024relukan} reconstructed bell-shaped basis functions using matrix operations and squared ReLU units, achieving up to a $20\times$ training speedup in four-layer networks over B-splines on synthetic benchmark functions. \textsc{ChebyKAN} \cite{ss2024chebykan} replaced B-splines with Chebyshev polynomial bases, achieving 98\% accuracy on MNIST with a compact parameterization, though without reporting comparative inference latency against MLPs. \textsc{TaylorKAN} \cite{chen2024taylorkan} applied Taylor series expansions as edge activations, demonstrating competitive results on image sharpness evaluation when combined with principal component analysis.

Beyond these models, several orthogonal polynomial formulations have been introduced that replace the spline grid with a polynomial degree hyperparameter. \textsc{JacobiKAN} \cite{spacelearner2024jacobikan, XIONG2026109414} incorporates orthogonal Jacobi polynomial bases, a concept generalized by fKAN \cite{aghaei2024fkan} with trainable fractional-order parameters for curvature adaptation. \textsc{LegendreKAN} \cite{HU2026109388} uses classical Legendre orthogonal polynomial recurrences to construct activation curves with theoretically supported convergence properties. Similarly, \textsc{GramKAN} \cite{khochawongwat2024gramkan} utilizes Gram polynomials—natively designed for discrete data—which are approximated here via continuous recurrence for tabular features. In an independent comparative analysis, Yu et al.\ \cite{yu2024mlpkan} examined KANs versus MLPs across computer vision, natural language processing, and standard machine learning tasks, finding that KANs primarily excelled at symbolic mathematical representation while standard MLPs maintained superior accuracy and efficiency on broader classification tasks. To address the specific requirements of structured data, Eslamian et al.\ \cite{eslamian2025tabkan} proposed \textsc{TabKAN}, which integrates KAN layers with input feature gating to suppress uninformative tabular variables.

Collectively, the existing literature on KAN architectures presents four notable gaps: (i) evaluations rely predominantly on synthetic functions or image classification rather than structured tabular data; (ii) comparisons are rarely performed against properly tuned neural and tree baselines across multi-dataset cohorts; (iii) inference latency and parameter counts are seldom profiled simultaneously; and (iv) evaluations typically overlook class imbalance and clinical calibration metrics. Consequently, the practical performance of KANs on real-world medical tabular benchmarks warrants systematic empirical evaluation.

\subsection{Deep Learning on Tabular Data}
\label{subsec:rw_tabular}

Whether deep neural networks can reliably outperform tree-based ensembles on tabular data remains a central question in applied machine learning. Gradient-boosted decision trees, particularly \textsc{XGBoost} \cite{chen2016xgboost}, established a dominant baseline across industrial and scientific tabular benchmarks due to their native handling of sparse features, missing values, and uninformative variables. Shwartz-Ziv and Armon \cite{shwartz2021tabular} conducted an extensive benchmark across 11 tabular datasets with standardized Bayesian hyperparameter optimization, showing that tree ensembles consistently outperformed deep architectures (such as TabNet and NODE). They also noted that deep models frequently underperformed on datasets beyond those highlighted in their original publications. Grinsztajn et al.\ \cite{grinsztajn2022why} further examined this dynamic across 45 OpenML benchmarks, observing that deep models struggle with irregular decision boundaries, uninformative features, and non-rotationally invariant feature spaces. These challenges apply directly to KANs, whose univariate edge functions were originally conceived for smooth mathematical compositions. Within this landscape, the \textsc{StandardMLP} with Batch Normalization and Dropout \cite{hornik1989multilayer} serves as the canonical neural baseline, while the \textsc{Random Forest} \cite{breiman2001random} provides an established non-neural reference.

\subsection{Statistical Frameworks for Multi-Dataset Algorithm Comparison}
\label{subsec:rw_stats}

Comparing machine learning models across multiple datasets requires statistical methodologies that account for cross-dataset variability and cross-validation dependencies. In a foundational methodological study, Dem\v{s}ar \cite{demsar2006statistical} highlighted that standard parametric tests (such as paired $t$-tests and ANOVA) often violate distributional assumptions when applied to cross-validated classifier scores, leading to inflated Type I errors. As a robust non-parametric alternative, Dem\v{s}ar recommended the Wilcoxon signed-rank test for pairwise comparisons and the Friedman test with post-hoc rank corrections for multi-model comparisons. However, frequentist tests have an important epistemological limitation: failing to reject the null hypothesis ($p \geq 0.05$) indicates only an absence of evidence against equality, not affirmative evidence \emph{for} practical equivalence.

To overcome this limitation, Benavoli et al.\ \cite{benavoli2017time} proposed a Bayesian testing framework incorporating the Bayesian Correlated $t$-Test and a Region of Practical Equivalence (ROPE). By defining a clinically negligible difference interval $[-\delta, +\delta]$ and integrating the posterior probability density within this region, the Bayesian ROPE approach explicitly quantifies the probability of practical equivalence ($P(\text{ROPE})$). To account for the correlation between folds in cross-validation resulting from overlapping training splits, Corani and Benavoli \cite{corani2015bayesian} derived an analytical variance adjustment based on the test fraction $\rho = 1/k$. This Bayesian framework provides the statistical foundation for our evaluation.

\subsection{Predictive Uncertainty, Calibration, and Imbalance in Clinical AI}
\label{subsec:rw_clinical}

Deploying machine learning models in healthcare requires properties beyond raw overall accuracy. First, models must deliver calibrated probability estimates: predicted probabilities must correspond to observed event rates \cite{van2019calibration}. Poorly calibrated models that produce overconfident risk estimates can result in inappropriate clinical prioritization and patient harm. Moreover, probabilistic calibration should be routinely evaluated using proper scoring rules such as the Brier Score. Rajpurkar et al.\ \cite{rajpurkar2022ai} emphasized that model robustness, computational efficiency, and calibration are essential for safe clinical deployment.

Second, medical tabular datasets typically present pronounced class imbalance, with positive conditions (e.g., biopsy-confirmed malignancies, fetal distress) accounting for only a small fraction of cases. Standard receiver operating characteristic curves (AUROC) can provide an overly optimistic assessment in this setting because the large number of true negatives keeps the false positive rate low. By contrast, the Precision-Recall curve and its area (AUPRC) focus exclusively on the true positive class, making AUPRC particularly informative for evaluating detection quality in rare clinical conditions.

Finally, clinical deployment imposes strict latency and resource constraints. While techniques like Monte Carlo Dropout \cite{gal2016dropout} and Deep Ensembles \cite{lakshminarayanan2017simple} can enhance uncertainty estimation, they increase computational latency linearly with the number of passes or models \cite{abdar2021review}. Because point-of-care decision support systems may operate on local hospital hardware or mobile workstations, computational cost (inference latency and memory footprint) represents an essential operational constraint.

Our study combines these complementary perspectives: evaluating KAN architectures and a tuned \textsc{StandardMLP} baseline on real-world medical tabular datasets using threshold-free AUPRC discrimination, Brier Score calibration, Bayesian ROPE equivalence testing, and comprehensive computational profiling.

% ===========================================================
% SECTION 3: MATERIALS AND METHODS
% ===========================================================

\section{Materials and Methods}
\label{sec:methods}

\subsection{Datasets and Preprocessing}
\label{subsec:datasets}

Seven publicly available medical tabular datasets, collectively denoted $\mathcal{D} = \{(\mathbf{x}_i, y_i)\}_{i=1}^N$, were retrieved from the UCI Machine Learning Repository \cite{dua2019uci} and are summarized in Table~\ref{tab:datasets}. The benchmark spans seven distinct clinical domains: Breast Cancer Wisconsin \cite{wolberg1992breast} ($N=569$, $d=30$, binary); Pima Indians Diabetes \cite{smith1988using} ($N=768$, $d=8$, binary); Cleveland Heart Disease \cite{detrano1989international} ($N=303$, $d=13$, multiclass); Chronic Kidney Disease \cite{rubini2015chronic} ($N=400$, $d=24$, binary); Parkinson's Disease \cite{little2007exploiting} ($N=195$, $d=22$, binary); Cervical Cancer \cite{fernandes2017transfer} ($N=858$, $d=32$, binary); and Cardiotocography \cite{ayres2000sisporto} ($N=2{,}126$, $d=21$, multiclass).

These datasets reflect realistic challenges encountered in clinical data science, including varying sample sizes ($N \in [195, 2{,}126]$), differing feature dimensions ($d \in [8, 32]$), missing laboratory readings, and pronounced class imbalance. In particular, the Cervical Cancer dataset exhibits an acute class imbalance with only $6.4\%$ biopsy-positive cases ($55$ of $858$), and the Cardiotocography dataset contains an $8.3\%$ pathological minority class, presenting demanding test conditions for minority-class detection.

\begin{table}[!t]
\renewcommand{\arraystretch}{1.3}
\caption{Summary of Benchmark Medical Datasets}
\label{tab:datasets}
\centering
\resizebox{\columnwidth}{!}{
\begin{tabular}{lccccc}
\toprule
\textbf{Dataset} & \textbf{$N$} & \textbf{$d$} & \textbf{Minority (\%)} & \textbf{Task} & \textbf{Clinical Domain} \\
\midrule
Cervical Cancer \cite{fernandes2017transfer}  & 858   & 32 & 6.4\%  & Binary     & Gynaecology / Oncology \\
Cardiotocography \cite{ayres2000sisporto}     & 2,126 & 21 & 8.3\%  & Multiclass & Obstetrics \\
Parkinson's \cite{little2007exploiting}       & 195   & 22 & 24.6\% & Binary     & Neurology \\
Pima Diabetes \cite{smith1988using}           & 768   & 8  & 34.9\% & Binary     & Endocrinology \\
Breast Cancer \cite{wolberg1992breast}        & 569   & 30 & 37.3\% & Binary     & Oncology \\
Chronic Kidney \cite{rubini2015chronic}       & 400   & 24 & 37.5\% & Binary     & Nephrology \\
Heart Disease \cite{detrano1989international} & 303   & 13 & 45.9\% & Multiclass & Cardiology \\
\bottomrule
\end{tabular}
}

\vspace{1ex}
\raggedright \footnotesize \textit{Note:} $N$ -- number of samples; $d$ -- number of features.
\end{table}

All preprocessing steps were implemented inside a strict pipeline to prevent data leakage. Estimators were fitted exclusively on the training split of each cross-validation fold and applied to the validation split via stateless transformations. For datasets with missing entries, continuous attributes were imputed using the column-wise median, except for Chronic Kidney Disease and Cervical Cancer, where $K$-Nearest Neighbors ($K=5$) imputation was applied to handle structured missingness. Categorical features were imputed using the mode and one-hot encoded. Continuous variables were standardized to zero mean and unit variance.

% -----------------------------------------------------------
\subsection{Model Architectures}
\label{subsec:models}

We evaluated eleven model architectures across the benchmark suite:

\textbf{Neural Baseline.} The \textsc{StandardMLP} baseline comprises a multi-layer feedforward network with Batch Normalization, Dropout regularisation ($p=0.1$), and Gaussian Error Linear Unit (GELU) activations \cite{hornik1989multilayer}. The architecture follows a pyramidal hidden-layer structure $[h, \lfloor h/2 \rfloor]$.

\textbf{Tree-Based Baseline.} The \textsc{Random Forest} classifier \cite{breiman2001random} provides an established non-neural reference for tabular data, implemented via Scikit-learn \cite{pedregosa2011scikit}.

\textbf{KAN Architectures.} Nine Kolmogorov-Arnold Network variants were evaluated:
\begin{enumerate}
    \item \textsc{FastKAN} \cite{li2024fastkan}: Approximates third-order B-splines through Gaussian Radial Basis Functions to eliminate knot-search operations.
    \item \textsc{Wav-KAN} \cite{bozorgasl2024wavkan}: Employs continuous/discrete wavelet transforms as edge activations for multiscale representation.
    \item \textsc{ReLU-KAN} \cite{qiu2024relukan}: Replaces splines with combinations of matrix additions, dot products, and rectified linear units.
    \item \textsc{ChebyKAN} \cite{ss2024chebykan}: Uses truncated Chebyshev polynomial bases of degree $k$.
    \item \textsc{TaylorKAN} \cite{chen2024taylorkan}: Utilizes low-order Taylor polynomial series expansions on network edges.
    \item \textsc{JacobiKAN} \cite{spacelearner2024jacobikan}: Jacobi orthogonal polynomial bases of degree $k$ with fixed shape parameters $\alpha=\beta=1$ (inputs squashed to $[-1,1]$ via $\tanh$).
    \item \textsc{LegendreKAN} \cite{HU2026109388}: Legendre orthogonal polynomial bases of degree $k$ over $[-1,1]$ (inputs squashed via $\tanh$).
    \item \textsc{GramKAN} \cite{khochawongwat2024gramkan}: Gram polynomial bases of degree $k$. Following common deep-learning implementations, we evaluate the basis through a continuous three-term recurrence on $\tanh$-normalized inputs, which approximates the discrete Gram formulation for tabular features.
    \item \textsc{TabKAN} \cite{eslamian2025tabkan}: Integrates KAN layers with a dedicated soft feature-gating mechanism that dynamically weights informative tabular features.
\end{enumerate}

% -----------------------------------------------------------
\subsection{Training Protocol and Hyperparameter Optimisation}
\label{subsec:training}

Evaluation was conducted using a 3$\times$5 Repeated Stratified $K$-Fold Cross-Validation design, yielding 15 independent validation evaluations per model-dataset pair ($105$ evaluations per architecture across all seven datasets). Stratified partitioning ensured preserved class ratios across each fold.

To ensure fair comparison, all models underwent automated hyperparameter optimization (HPO) using Optuna \cite{akiba2019optuna} with the Tree-structured Parzen Estimator (TPE) algorithm. A budget of 15 trials was allocated per fold, with unpromising configurations pruned using Optuna's \texttt{MedianPruner}. HPO was conducted inside an inner stratified 80/20 train-validation split of the training fold data. The inner validation loss guided parameter selection, ensuring no information from the outer test fold leaked into the tuning process.

The search space for neural models included the learning rate ($\text{LogUniform}[10^{-4}, 10^{-2}]$), weight decay ($\text{LogUniform}[10^{-6}, 10^{-3}]$), and hidden dimension $h \in \{32, 64, 128\}$. Architecture-specific parameters included \texttt{grid\_size} $\in \{3, 5, 10\}$ for spline models, polynomial degree $\in \{3, 4, 6\}$ for polynomial KANs, and \texttt{num\_grids} $\in \{4, 8, 12\}$ for FastKAN and TabKAN. For the Random Forest, estimators were sampled from $[50, 300]$, maximum depth from $[3, 15]$, and minimum split samples from $[2, 10]$.

Neural models were trained using AdamW with early stopping (patience of 10 epochs, minimum validation loss improvement $\delta = 10^{-4}$). Binary tasks were optimized using binary cross-entropy with logits loss, and multiclass tasks using categorical cross-entropy.

% -----------------------------------------------------------
\subsection{Evaluation Metrics}
\label{subsec:metrics}

Models were evaluated along two primary dimensions: predictive performance and computational efficiency.

\subsubsection{Discriminative Performance and AUPRC}

Given the prominent class imbalance across the benchmark datasets, the Area Under the Precision-Recall Curve (AUPRC) serves as our primary discriminative metric. In medical contexts with low disease prevalence, AUROC can yield misleadingly high scores because the large quantity of true negatives inflates the denominator of the false positive rate ($FP / (FP + TN)$) \cite{saito2015precision}. In contrast, AUPRC focuses exclusively on the positive minority class, capturing the trade-off between positive predictive value (precision) and sensitivity (recall):
\begin{equation*}
    \text{AUPRC} = \sum_{n} (R_n - R_{n-1}) P_n,
\end{equation*}
where $P_n$ and $R_n$ represent the precision and recall values at the $n$-th classification threshold. For multiclass tasks, macro-averaged One-vs-Rest (OvR) binarization was applied across all classes.

Supplementary discrimination metrics include the Matthews Correlation Coefficient (MCC) \cite{chicco2020advantages}, AUROC, macro-averaged F1-Score, and Balanced Accuracy.

\subsubsection{Calibration}

Probabilistic calibration was assessed using the Brier Score \cite{van2019calibration}, measuring the mean squared difference between predicted probabilities $\hat{p}_i$ and observed binary outcomes $y_i \in \{0, 1\}$:
\begin{equation*}
    \text{BS} = \frac{1}{N} \sum_{i=1}^{N} \left( \hat{p}_i - y_i \right)^2.
\end{equation*}
For multiclass datasets, the multi-class formulation based on one-vs-all decomposition was applied \cite{ferro2012bias}. Lower Brier scores indicate better calibration.

\subsubsection{Computational Efficiency}

Three computational metrics were recorded: (i) per-sample inference latency ($t_\text{inf}$, in milliseconds), calculated by dividing the wall-clock evaluation time of the validation fold by the sample count; (ii) trainable parameter count ($|\theta|$); and (iii) total training time ($T_\text{train}$, in seconds per fold).

% -----------------------------------------------------------
\subsection{Statistical Analysis Framework}
\label{subsec:stats}

To differentiate genuine architectural differences from sampling variance across datasets, we employed both frequentist and Bayesian statistical tests \cite{demsar2006statistical,benavoli2017time}.

\subsubsection{Frequentist Testing}

Dataset-level performance scores (averaged across the $15$ validation folds per dataset) were compared pairwise against the \textsc{StandardMLP} baseline using the two-tailed non-parametric Wilcoxon signed-rank test \cite{demsar2006statistical}. To control the family-wise error rate across multiple architecture comparisons, the Holm-Bonferroni step-down correction was applied at $\alpha = 0.05$. Omnibus multi-model comparisons across the benchmark were evaluated using the Friedman test on dataset-level model ranks ($1 = \text{best}$ to $k = 11 = \text{worst}$), followed by the Nemenyi post-hoc test. Cross-dataset average ranks and pairwise significance thresholds were visualized via Critical Difference (CD) diagrams, where models whose average rank difference is smaller than $\text{CD} = q_{\alpha}\sqrt{k(k+1)/(6N)}$ are connected by a horizontal clique bar \cite{demsar2006statistical}.

\subsubsection{Bayesian Correlated $t$-Test with ROPE}

Because non-significant frequentist $p$-values cannot formally establish equivalence, we implemented the Bayesian Correlated $t$-Test with a Region of Practical Equivalence (ROPE) \cite{corani2015bayesian,benavoli2017time}. For models $A$ and $B$, paired fold differences $\Delta_j = s^{(A)}_j - s^{(B)}_j$ across all $n=15$ folds were evaluated. To account for the correlation between folds sharing training data in cross-validation, the posterior variance was adjusted using:
\begin{equation*}
    \hat{\sigma}_\text{corr}^2 = \left( \frac{1}{n} + \frac{\rho}{1-\rho} \right) \hat{\sigma}_{\Delta}^2, \quad \rho = \frac{1}{k} = 0.2,
\end{equation*}
where $\rho$ represents the test-to-training ratio for $k=5$ folds \cite{corani2015bayesian}.

A clinically motivated ROPE interval of $\pm 0.01$ absolute units ($1\%$) was established \cite{benavoli2017time}. Integrating the posterior Student's $t$-distribution over this interval yields three mutually exclusive probabilities:
\begin{align*}
    P(A \succ B) &= P(\Delta > 0.01), \\
    P(B \succ A) &= P(\Delta < -0.01), \\
    P(\text{ROPE}) &= P(|\Delta| \leq 0.01).
\end{align*}
When $P(\text{ROPE}) \geq 0.90$, practical equivalence between architectures is considered supported by the evidence.

% -----------------------------------------------------------
\subsection{Hardware and Execution Environment}
\label{subsec:hardware}

To ensure reproducible computational latency profiling, all benchmarks were executed locally in a dedicated environment on an Apple MacBook Pro equipped with an Apple M5 Pro system-on-chip with 16-core CPU, high-performance GPU, and unified memory architecture. Neural network operations were accelerated using Apple's Metal Performance Shaders (MPS) backend within PyTorch 2.0+ \cite{paszke2019pytorch}. Consistent seeds (random state 42 for data partitioning and initialization) were enforced across all trials. Inference latency was recorded using high-resolution monotonic clocks over warm forward passes to prevent cold-start artifacts. Numerical processing and data manipulation relied on NumPy \cite{harris2020array}, pandas \cite{mckinney2010pandas}, and SciPy \cite{virtanen2020scipy}, while all figures were produced using Matplotlib \cite{hunter2007matplotlib} and Seaborn \cite{waskom2021seaborn}. 
% ===========================================================
% SECTION 4: RESULTS
% ===========================================================

\section{Results}
\label{sec:results}

The following section presents the empirical findings from the 3$\times$5 Repeated Stratified $K$-Fold benchmark across all seven medical datasets and eleven model architectures ($N = 105$ evaluations per architecture). Results are reported as mean $\pm$ standard deviation aggregated across all 15 independent folds per dataset. Unless stated otherwise, cross-dataset comparative summaries refer to grand means computed across the full benchmark.

\begin{figure*}[!t]
    \centering
    \includegraphics[width=\textwidth]{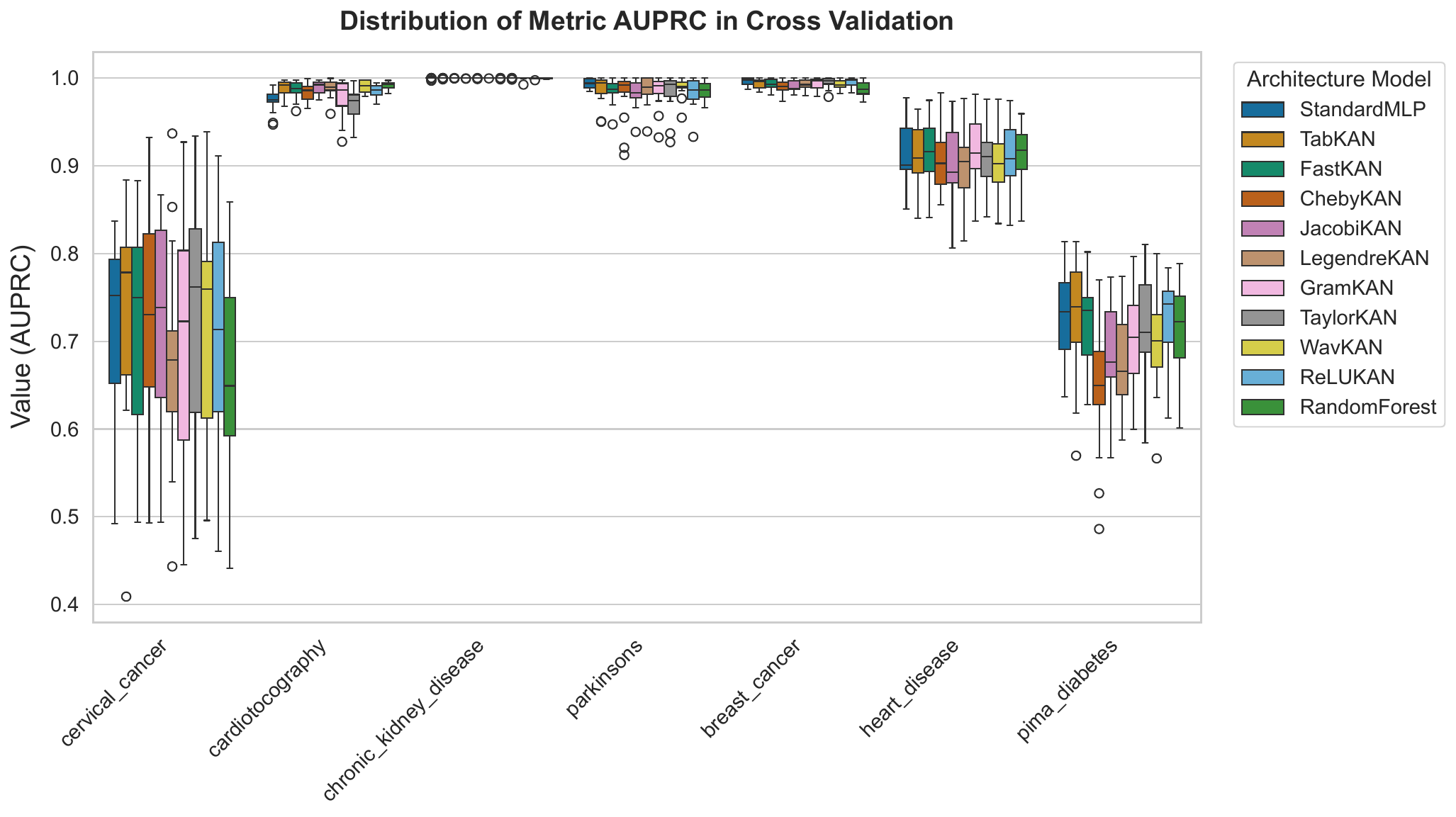}
    \caption{Distribution of per-fold Area Under the Precision-Recall Curve (AUPRC) scores for all eleven evaluated architectures across the seven medical datasets, shown as box-and-whisker plots (median, interquartile range, and 1.5-IQR whiskers). The results illustrate that while overall performance distributions overlap across balanced datasets, dedicated architectures such as TabKAN and FastKAN demonstrate competitive or superior retention of discriminative capacity on imbalanced cohorts (e.g., Cervical Cancer and Cardiotocography).}
    \label{fig:boxplot_auprc}
\end{figure*}

% -----------------------------------------------------------
\subsection{Predictive Performance Under Class Imbalance}
\label{subsec:performance}

\subsubsection{Discrimination Metrics}

Across all seven medical datasets ($N=105$ folds), \textsc{TabKAN} achieved the highest grand mean AUPRC (see Table~\ref{tab:grand_summary_auprc}) at $0.9061 \pm 0.1262$ with an average rank of $3.43$, followed by the baseline \textsc{StandardMLP} ($0.9028 \pm 0.1278$, average rank $4.43$), \textsc{FastKAN} ($0.9013 \pm 0.1323$), \textsc{ReLUKAN} ($0.8999 \pm 0.1358$), \textsc{TaylorKAN} ($0.8992 \pm 0.1313$), and \textsc{WavKAN} ($0.8980 \pm 0.1384$). Polynomial KANs recorded grand mean AUPRC values between $0.8894$ (\textsc{LegendreKAN}) and $0.8962$ (\textsc{GramKAN}), while the \textsc{RandomForest} baseline attained $0.8923 \pm 0.1455$. Note that the high variability across datasets results in a large standard deviation for AUPRC but it concerns also the other metrics analyzed (Table~\ref{tab:grand_summary_auprc}).

The slight advantage of dedicated KAN architectures was most evident on datasets characterized by pronounced minority-class prevalence:
\begin{itemize}
    \item \textit{Cervical Cancer} ($6.4\%$ positive prevalence): \textsc{TabKAN} attained an AUPRC of $0.7346 \pm 0.0891$, outperforming \textsc{StandardMLP} ($0.7150 \pm 0.0880$) and the \textsc{RandomForest} baseline ($0.6609 \pm 0.0849$). Furthermore, \textsc{TabKAN} achieved a precision of $79.4\%$ on biopsy-positive cases, compared to $66.5\%$ for \textsc{StandardMLP}.
    \item \textit{Cardiotocography} ($8.3\%$ minority class): \textsc{WavKAN} and \textsc{TabKAN} attained mean AUPRC scores of $0.9902 \pm 0.0091$ and $0.9884 \pm 0.0073$, respectively, exceeding \textsc{StandardMLP} ($0.9741 \pm 0.0194$) and achieving superior minority recall ($96.9\%$ and $95.6\%$ vs.\ $92.7\%$).
    \item \textit{Pima Diabetes ($34.9\%$ positive prevalence)}: \textsc{TabKAN} reached an AUPRC of $0.7279 \pm 0.0632$, performing slightly ahead of \textsc{StandardMLP} ($0.7263 \pm 0.0652$) and \textsc{ReLUKAN} ($0.7240 \pm 0.0617$).
\end{itemize}

\begin{figure}[!t]
    \centering
    
    \begin{subfigure}[b]{\columnwidth}
        \centering
        \includegraphics[width=\textwidth]{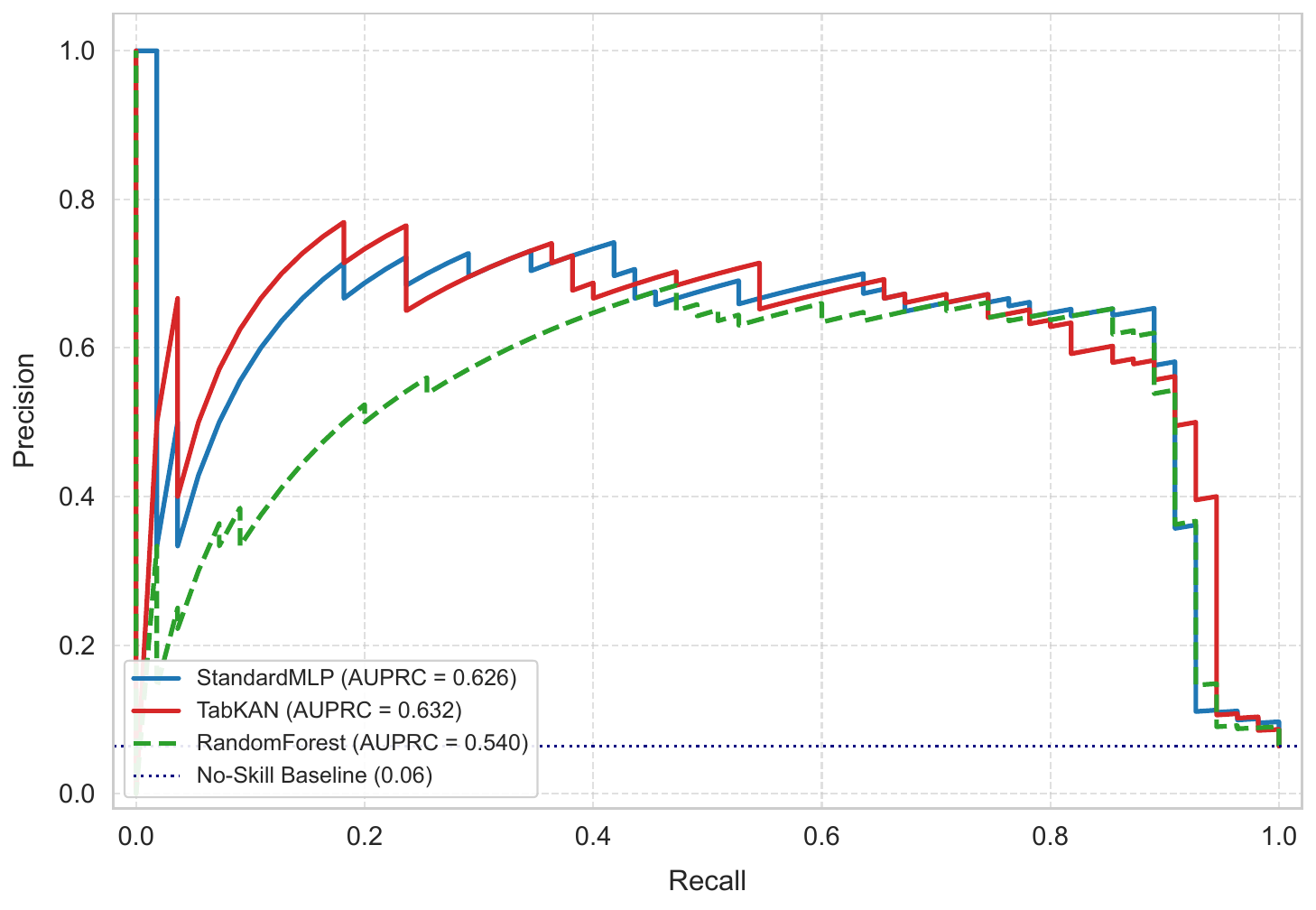}
        \caption{Cervical Cancer ($6.4\%$ pos.)}
        \label{fig:pr_cervical}
    \end{subfigure}
    
    \vspace{1.5em}
    
    \begin{subfigure}[b]{\columnwidth}
        \centering
        \includegraphics[width=\textwidth]{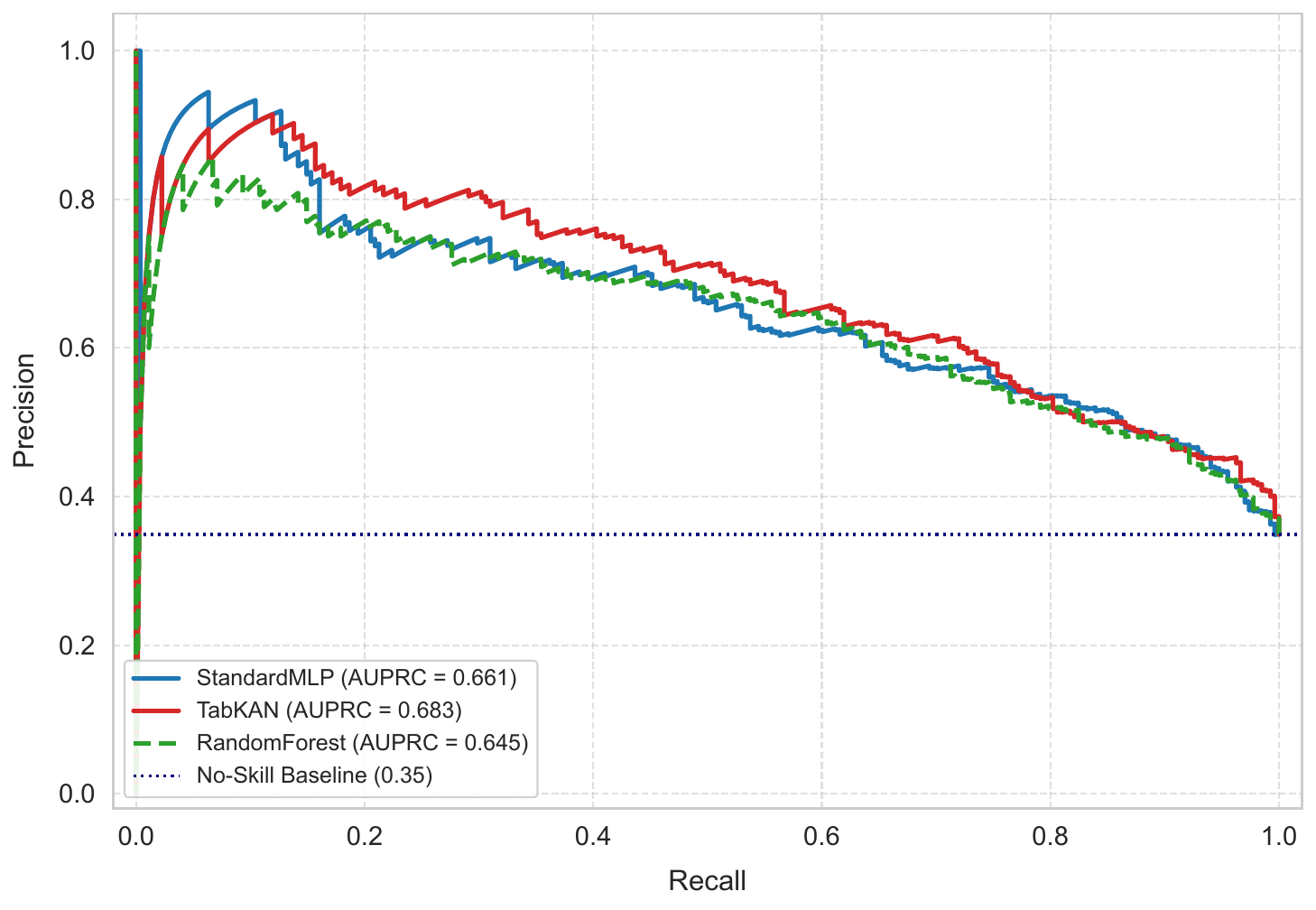}
        \caption{Pima Diabetes ($34.9\%$ pos.)}
        \label{fig:pr_pima}
    \end{subfigure}
    
    \caption{Precision-Recall curves for representative clinical cohorts with high and moderate class imbalance, comparing neural architectures and Random Forest baseline across cross-validation folds.}
    \label{fig:pr_curves}
\end{figure}

Precision-Recall trajectories for two representative cohorts with high (Fig.~\ref{fig:pr_cervical}) and moderate (Fig.~\ref{fig:pr_pima}) class imbalance are shown in Fig.~\ref{fig:pr_curves}. In both settings, \textsc{TabKAN} and related KAN variants demonstrate higher precision across intermediate-to-high recall ranges compared to standard MLPs and tree ensembles.

\begin{figure*}[!t]
    \centering
    \includegraphics[width=0.85\textwidth]{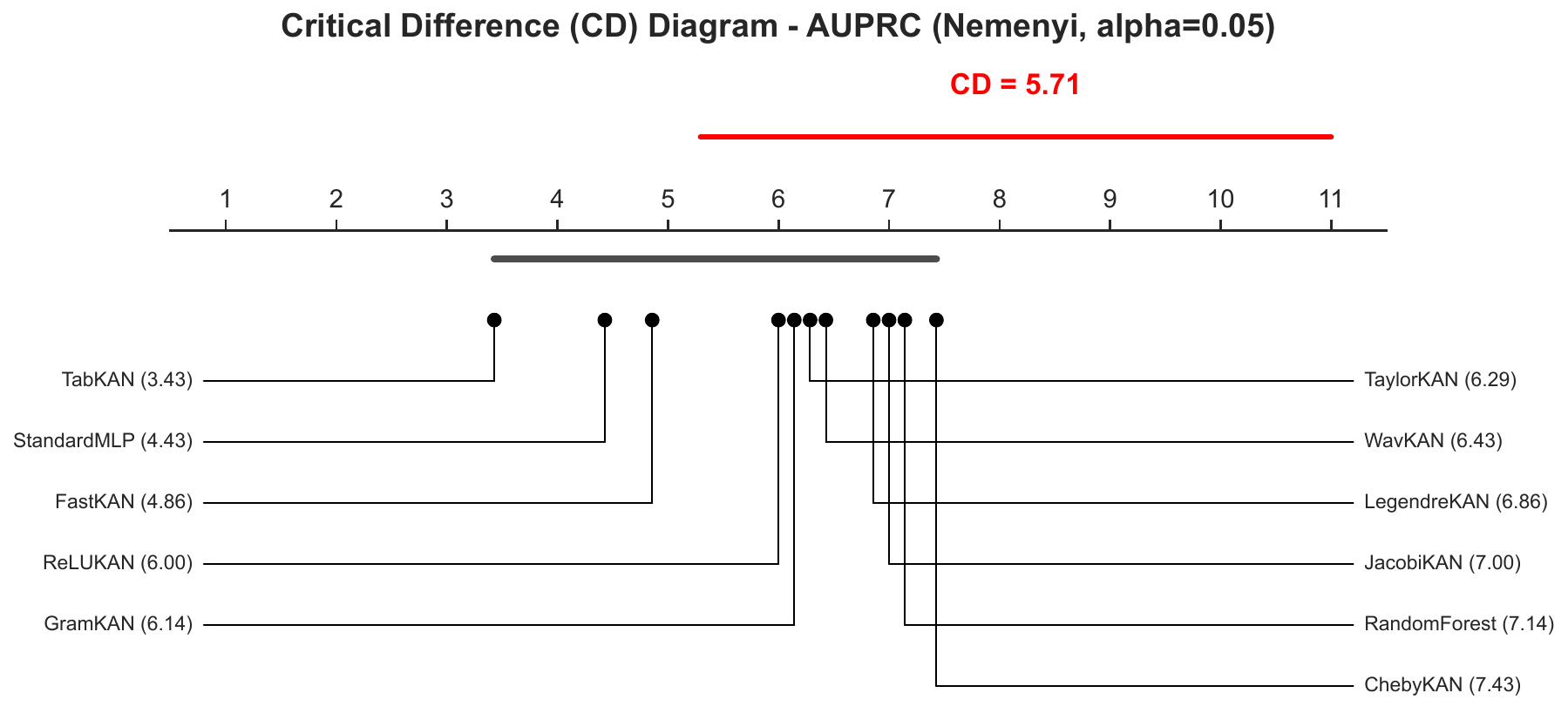}
    \caption{Critical Difference (CD) Diagram for AUPRC constructed from average ranks across all seven benchmark datasets under the Nemenyi post-hoc test ($\alpha = 0.05$). The horizontal axis represents the average rank of each model, where a lower rank indicates higher AUPRC. To interpret the diagram: the solid red line at the top visualizes the required critical difference distance ($CD = 5.71$), and the thick gray line connects models whose pairwise rank differences are smaller than this threshold (i.e., they are not statistically significantly different). While \textsc{TabKAN} achieves the lowest average rank ($3.43$), the critical difference connects all evaluated models in a single clique. This indicates that rank differences across the heterogeneous benchmark do not cross the family-wise significance threshold, rendering all architectures statistically equivalent under this frequentist test.}
    \label{fig:cd_diagram_auprc}
\end{figure*}

The Critical Difference (CD) Diagram for AUPRC is shown in Fig.~\ref{fig:cd_diagram_auprc}. While \textsc{TabKAN} obtained the lowest average rank across the seven datasets ($3.43$), the critical difference threshold ($CD = 5.71$) connects the evaluated models within a shared clique under the conservative Nemenyi post-hoc test ($\alpha = 0.05$). This indicates that, when averaged across highly diverse clinical domains, rank differences alone do not achieve family-wise statistical separation.

\subsubsection{Frequentist and Bayesian Statistical Analysis}

Pairwise Wilcoxon signed-rank tests comparing \textsc{StandardMLP} against the best-performing KAN variants across the seven datasets yielded unadjusted $p$-values between $0.109$ and $0.813$ for AUPRC. After Holm-Bonferroni correction, no architecture achieved formal statistical significance at $\alpha = 0.05$. As noted in Section~\ref{subsec:stats}, non-parametric tests have limited power when aggregating across a small number of dataset-level observations ($n = 7$), where the minimum achievable two-sided $p$-value cannot clear conservative step-down thresholds.

\begin{table}[!t]
\renewcommand{\arraystretch}{1.3}
\caption{Bayesian Correlated $t$-Test Results (AUPRC). ROPE interval $= \pm 0.01$.}
\label{tab:rope_auprc}
\centering
\resizebox{\columnwidth}{!}{
\begin{tabular}{lccccc}
\toprule
\textbf{Dataset} & \textbf{Best KAN} & $P(\text{MLP} \succ)$ & $P(\text{KAN} \succ)$ & $P(\text{ROPE})$ & $\bar{\Delta}$ \\
\midrule
Breast Cancer             & TaylorKAN    & <0.001   & <0.001   & 1.000    & +0.0014 \\
Cardiotocography          & WavKAN       & <0.001   & 0.885    & 0.115    & -0.0161 \\
Chronic Kidney Disease    & LegendreKAN  & <0.001   & <0.001   & 1.000    & -0.0004 \\
Heart Disease             & GramKAN      & 0.081    & 0.116    & 0.803    & -0.0008 \\
Parkinson's Disease       & WavKAN       & 0.130    & 0.005    & 0.865    & +0.0043 \\
Pima Diabetes             & TabKAN       & 0.289    & 0.343    & 0.368    & -0.0016 \\
Cervical Cancer           & TabKAN       & 0.223    & 0.598    & 0.179    & -0.0195 \\
\midrule
\textbf{OVERALL (Pooled)} & \textit{Per-Dataset Best} & 0.169 & 0.364 & 0.467 & -0.0047 \\
\bottomrule
\end{tabular}
}

\vspace{1ex}
\raggedright \footnotesize \textit{Note:} The representative KAN variant for each dataset was identified based on the highest mean Area Under the Precision-Recall Curve (AUPRC) across all 15 cross-validation folds. Acronyms and notation: ROPE = Region of Practical Equivalence (margin $[-0.01, +0.01]$); $P(\text{MLP} \succ) = P(\Delta > 0.01)$ denotes posterior probability that StandardMLP outperforms KAN by more than the practical equivalence margin; $P(\text{KAN} \succ) = P(\Delta < -0.01)$ denotes posterior probability that KAN outperforms StandardMLP; $P(\text{ROPE}) = P(|\Delta| \le 0.01)$ denotes posterior probability of practical equivalence; $\bar{\Delta} = \text{Posterior Mean Difference } (\text{StandardMLP} - \text{KAN})$, where negative values indicate higher AUPRC for KAN.
\end{table}

The Bayesian Correlated $t$-Test with ROPE \cite{corani2015bayesian,benavoli2017time} provides a more nuanced probabilistic characterization of these differences, summarized in Table~\ref{tab:rope_auprc}. On datasets with balanced or moderately balanced classes, the posterior probability of practical equivalence reaches or approaches unity: $P(\text{ROPE}) = 1.000$ on \textit{Breast Cancer} and \textit{Chronic Kidney Disease}, $0.803$ on \textit{Heart Disease}, and $0.865$ on \textit{Parkinson's Disease}. In these contexts, KAN variants and \textsc{StandardMLP} exhibit practically indistinguishable discriminative performance.

Conversely, on datasets with severe class imbalance, the Bayesian posterior shifts toward KAN superiority. On \textit{Cardiotocography}, \textsc{WavKAN} records $P(\text{KAN} \succ) = 0.885$ with $P(\text{ROPE}) = 0.115$ and a posterior mean difference of $\bar{\Delta} = -0.0161$, providing strong Bayesian evidence of higher AUPRC than \textsc{StandardMLP}. On \textit{Cervical Cancer}, \textsc{TabKAN} achieves $P(\text{KAN} \succ) = 0.598$ versus $P(\text{MLP} \succ) = 0.223$ ($P(\text{ROPE}) = 0.179$, $\bar{\Delta} = -0.0195$), indicating a substantial probability of clinically meaningful improvement on rare positive cases. In the pooled cross-dataset analysis ($N=105$), the probability of KAN superiority exceeds that of the MLP ($P(\text{KAN} \succ) = 0.364$ vs.\ $P(\text{MLP} \succ) = 0.169$), with $P(\text{ROPE}) = 0.467$.

These findings suggest that while KANs and MLPs display comparable predictive capacity in balanced settings, specialized KAN architectures (particularly \textsc{TabKAN} and \textsc{WavKAN}) can offer meaningful discriminative benefits when detecting rare pathological conditions.

% -----------------------------------------------------------
\subsection{Model Calibration}
\label{subsec:calibration}

\begin{figure*}[!t]
    \centering
    \includegraphics[width=\textwidth]{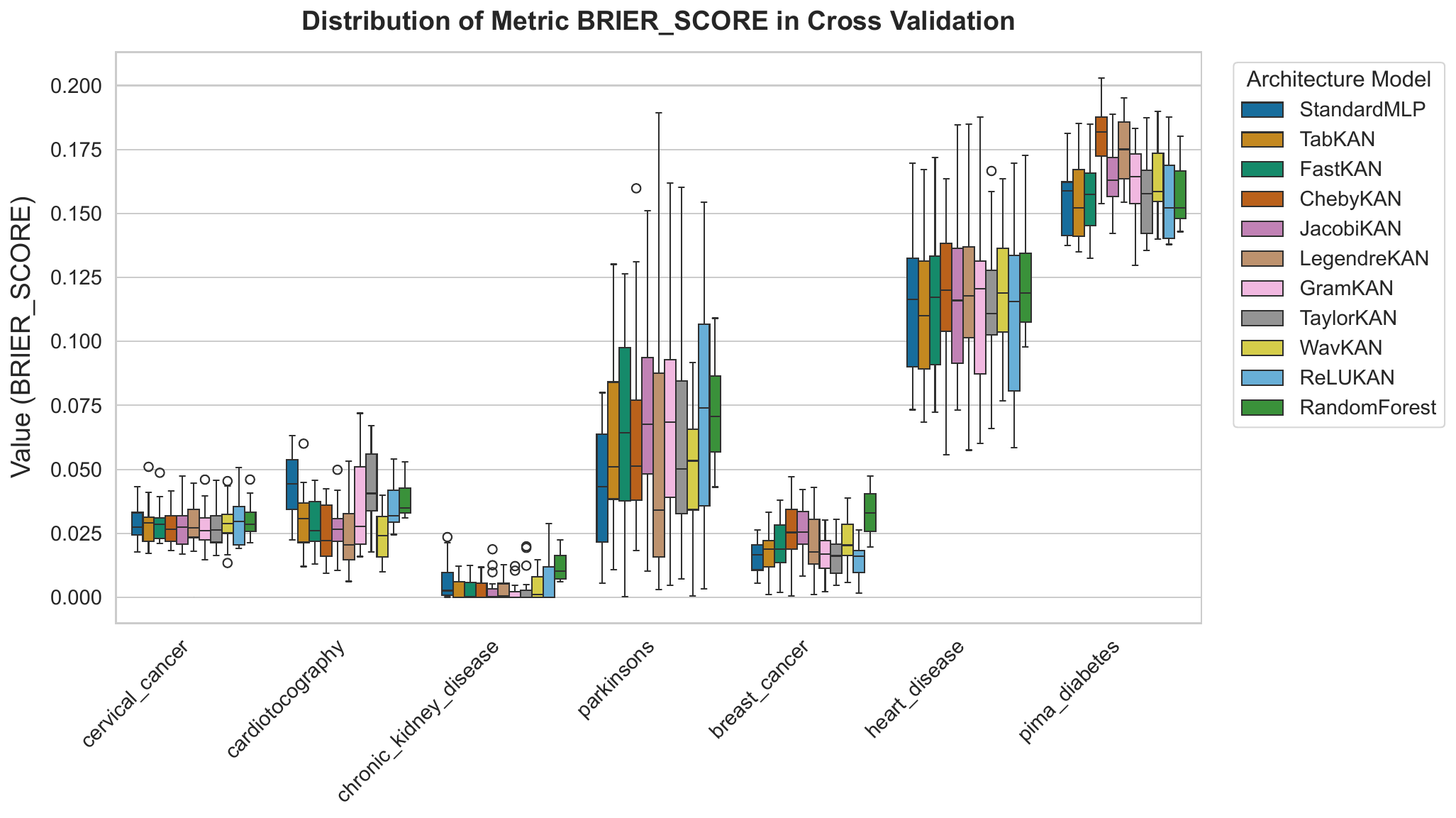}
    \caption{Distribution of Brier Scores across the repeated cross-validation folds. Lower scores indicate better probabilistic calibration, a critical requirement for safe clinical risk stratification.}
    \label{fig:brier_score}
\end{figure*}

Safe clinical integration requires not only accurate classification, but also reliable probability estimates \cite{van2019calibration}. Model calibration was evaluated through the Brier Score, where lower scores indicate superior probabilistic fidelity. Distributional profiles across cross-validation folds are presented in Fig.~\ref{fig:brier_score}, and grand summary scores are provided in Table~\ref{tab:brier}.

\begin{table}[!t]
\renewcommand{\arraystretch}{1.3}
\caption{Grand Mean Brier Score across all datasets and folds ($\text{Grand Mean} \pm \text{Std}$, lower is better).}
\label{tab:brier}
\centering
\begin{tabular}{lc}
\toprule
\textbf{Model} & \textbf{Grand Mean Brier Score} \\
\midrule
TabKAN        & $0.0580 \pm 0.0547$ \\
StandardMLP   & $0.0585 \pm 0.0542$ \\
WavKAN        & $0.0591 \pm 0.0569$ \\
FastKAN       & $0.0600 \pm 0.0559$ \\
ReLUKAN       & $0.0606 \pm 0.0558$ \\
TaylorKAN     & $0.0606 \pm 0.0560$ \\
LegendreKAN   & $0.0611 \pm 0.0628$ \\
GramKAN       & $0.0615 \pm 0.0589$ \\
JacobiKAN     & $0.0624 \pm 0.0583$ \\
ChebyKAN      & $0.0634 \pm 0.0628$ \\
RandomForest  & $0.0665 \pm 0.0523$ \\
\bottomrule
\end{tabular}

\vspace{1ex}
\raggedright \footnotesize \textit{Note:} BS = Brier Score (quadratic calibration loss measuring mean squared difference between predicted class probabilities and binary ground-truth labels; ranges from $0$ to $1$, where lower scores indicate better probabilistic calibration). Values represent grand mean $\pm$ standard deviation aggregated across all seven clinical datasets ($N=105$ evaluations per architecture).
\end{table}

\textsc{TabKAN} achieved the lowest grand mean Brier Score ($0.0580 \pm 0.0547$), closely matched by \textsc{StandardMLP} ($0.0585 \pm 0.0542$) and \textsc{WavKAN} ($0.0591 \pm 0.0569$). Spline- and wavelet-based KANs maintained competitive calibration alongside the neural baseline, whereas polynomial variants exhibited marginally higher Brier scores ($0.0611$--$0.0634$). \textsc{RandomForest} recorded the highest overall calibration error ($0.0665 \pm 0.0523$), reflecting the known tendency of uncalibrated decision trees to produce stepped, miscalibrated posterior estimates.

Bayesian ROPE analysis on Brier scores confirms practical equivalence between leading KAN variants and the MLP baseline across most datasets: posterior probabilities $P(\text{ROPE})$ exceeded $0.73$ on five of the seven datasets (reaching $0.999$ on Cervical Cancer and Pima Diabetes). Overall, these data indicate that modern KAN implementations can achieve calibration quality comparable to tuned MLPs without requiring secondary post-hoc calibration procedures.

% -----------------------------------------------------------
\subsection{Computational Cost and Efficiency Trade-offs}
\label{subsec:compute}

While predictive discrimination and calibration are closely matched or slightly in favor of specific KAN variants on imbalanced tasks, computational profiling reveals systematic differences in inference latency, parameter footprint, and training duration.

\begin{figure}[!t]
    \centering
    \includegraphics[width=\columnwidth]{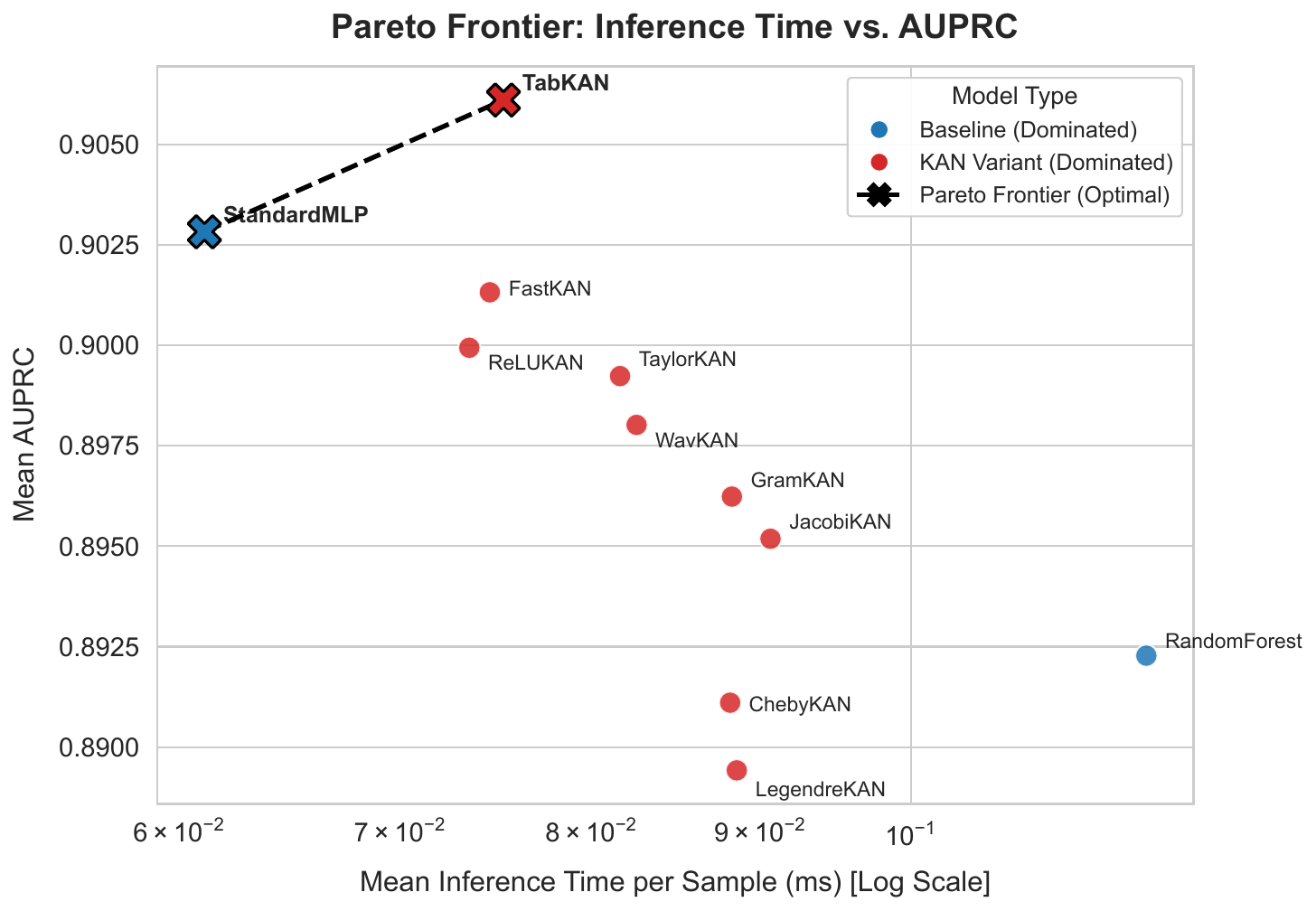}
    \caption{Pareto Frontier of grand mean AUPRC versus mean per-sample inference latency (log scale) across all eleven evaluated architectures. The empirical Pareto frontier is formed by \textsc{StandardMLP} (lowest latency: $0.0611$\,ms) and \textsc{TabKAN} (highest AUPRC: $0.9061$). Other KAN variants and baseline models remain within the dominated region, highlighting the operational trade-off between baseline execution speed and KAN minority-class retrieval performance.}
    \label{fig:pareto}
\end{figure}

\subsubsection{Inference Latency}

Per-sample inference latencies measured on dedicated Apple Silicon hardware (MPS backend) are reported in Table~\ref{tab:inference}. \textsc{StandardMLP} achieved the lowest median latency at $\bar{t}_\text{inf} = 0.0611 \pm 0.0165$\,ms per sample. In comparison, KAN architectures introduced measurable latency overheads: \textsc{ReLUKAN} was $1.20\times$ slower ($0.0735 \pm 0.0256$\,ms), \textsc{FastKAN} was $1.22\times$ slower ($0.0745 \pm 0.0244$\,ms), and \textsc{TabKAN} was $1.23\times$ slower ($0.0752 \pm 0.0260$\,ms). Polynomial and wavelet variants imposed larger overheads, reaching $1.34\times$ for \textsc{TaylorKAN} ($0.0816$\,ms), $1.35\times$ for \textsc{WavKAN} ($0.0826$\,ms), and up to $1.48\times$ for \textsc{JacobiKAN} ($0.0906 \pm 0.0298$\,ms). The tree-based \textsc{RandomForest} recorded an inference latency of $0.1178 \pm 0.0658$\,ms ($1.93\times$ the MLP baseline).

\begin{table}[!t]
\renewcommand{\arraystretch}{1.3}
\caption{Computational Cost Summary (Grand Mean across all datasets and folds).}
\label{tab:inference}
\centering
\resizebox{\columnwidth}{!}{
\begin{tabular}{lcccc}
\toprule
\textbf{Model} & \textbf{$\bar{t}_\text{inf}$ (ms)} & \textbf{Ratio vs.\ MLP} & \textbf{$|\theta|$ (Grand Mean)} & \textbf{$\bar{T}_\text{train}$ (s)} \\
\midrule
StandardMLP  & 0.0611 & $1.00\times$ & 8,273   & 1.22 \\
ReLUKAN      & 0.0735 & $1.20\times$ & 73,906  & 2.04 \\
FastKAN      & 0.0745 & $1.22\times$ & 39,143  & 1.66 \\
TabKAN       & 0.0752 & $1.23\times$ & 56,990  & 2.04 \\
TaylorKAN    & 0.0816 & $1.34\times$ & 23,885  & 1.86 \\
WavKAN       & 0.0826 & $1.35\times$ & 103,347 & 2.23 \\
ChebyKAN     & 0.0881 & $1.44\times$ & 23,361  & 1.64 \\
GramKAN      & 0.0882 & $1.44\times$ & 23,678  & 1.86 \\
LegendreKAN  & 0.0885 & $1.45\times$ & 23,622  & 1.73 \\
JacobiKAN    & 0.0906 & $1.48\times$ & 22,892  & 1.61 \\
RandomForest & 0.1178 & $1.93\times$ & ---     & 0.07 \\
\bottomrule
\end{tabular}
}

\vspace{1ex}
\raggedright \footnotesize \textit{Note:} Computational metrics: $\bar{t}_\text{inf}$ = Mean per-sample inference latency in milliseconds (ms), measured with batch size 1 on Apple Silicon MPS hardware; Ratio vs.\ MLP = Relative inference latency factor normalized to the StandardMLP baseline ($1.00\times$); $|\theta|$ = Mean count of trainable network parameters; $\bar{T}_\text{train}$ = Mean wall-clock training duration per cross-validation fold in seconds (s). Values denote grand means across all 7 benchmark datasets ($N=105$ cross-validation evaluations per architecture).
\end{table}

\subsubsection{Parameter Complexity}

The \textsc{StandardMLP} proved notably more parameter-efficient than all evaluated KAN models. With an average trainable parameter count of $8{,}273 \pm 5{,}210$, the MLP required between $2.8\times$ and $12.5\times$ fewer parameters than the KAN family. Polynomial KAN variants ranged from $22{,}892$ (\textsc{JacobiKAN}) to $23{,}885$ (\textsc{TaylorKAN}), while spline and wavelet models carried larger parameter footprints: $39{,}143$ (\textsc{FastKAN}), $56{,}990$ (\textsc{TabKAN}), $73{,}906$ (\textsc{ReLUKAN}), and $103{,}347$ (\textsc{WavKAN}). This expansion reflects the structural mechanism of KANs, which substitute scalar synaptic weights with parameter vectors for edge-based activation bases.

\subsubsection{Training Duration}

Per-fold training times were consistent with the architectural complexity patterns. The \textsc{StandardMLP} completed training in an average of $1.22 \pm 1.12$\,s per fold, whereas KAN variants required between $1.61$\,s (\textsc{JacobiKAN}) and $2.23$\,s (\textsc{WavKAN}), with \textsc{TabKAN} averaging $2.04 \pm 1.51$\,s. While non-neural \textsc{RandomForest} trained significantly faster ($0.07 \pm 0.05$\,s), the training overhead of KANs remained manageable within standard automated machine learning pipelines.

\subsubsection{Pareto Frontier Analysis}

The operational trade-off between discriminative quality (AUPRC) and execution latency is visualised in Fig.~\ref{fig:pareto}. Unlike single-metric evaluations that might suggest unconditional dominance, the empirical Pareto frontier is defined jointly by two models:
\begin{enumerate}
    \item \textsc{StandardMLP}: Establishes the lower bound on inference latency ($0.0611$\,ms) while maintaining high discriminative utility ($0.9028$ AUPRC).
    \item \textsc{TabKAN}: Achieves the highest overall discriminative performance ($0.9061$ AUPRC) at a moderate latency cost ($0.0752$\,ms, a $1.23\times$ overhead).
\end{enumerate}
All other architectures occupy strictly interior (dominated) positions within the trade-off space. In practical terms, this result suggests that selection between an MLP and a tailored KAN depends on clinical deployment constraints: when latency is critical (such as bedside monitoring or edge diagnostics), \textsc{StandardMLP} remains optimal; when maximizing sensitivity and precision on imbalanced disease classes takes priority, \textsc{TabKAN} provides a viable, higher-performing alternative.

% ===========================================================
% SECTION 5: CONCLUSION
% ===========================================================

\section{Conclusion}
\label{sec:conclusion}

This study provided a standardized benchmark comparing Kolmogorov-Arnold Networks (KANs), Multi-Layer Perceptrons (MLPs), and tree-based baselines across seven medical tabular datasets under repeated stratified cross-validation with nested Bayesian hyperparameter optimization. Motivated by the prevalence of class imbalance in clinical decision-making, the evaluation centered on the Area Under the Precision-Recall Curve (AUPRC), probabilistic calibration via the Brier Score, and computational efficiency, supplemented by formal Bayesian ROPE testing.

Our findings lead to three main conclusions regarding the practical utility of KANs in medical tabular domains:

First, on datasets with severe class imbalance, specialized KAN architectures demonstrate tangible discriminative advantages. In particular, \textsc{TabKAN} attained the highest grand mean AUPRC across the benchmark ($0.9061 \pm 0.1262$, rank $3.43$), outperforming \textsc{StandardMLP} on cohorts such as Cervical Cancer ($0.7346$ vs.\ $0.7150$, with precision reaching $79.4\%$ compared to $66.5\%$) and Cardiotocography ($0.9884$ vs.\ $0.9741$). On these imbalanced tasks, Bayesian equivalence tests indicated elevated posterior probabilities of KAN superiority ($P(\text{KAN} \succ) \in [0.598, 0.885]$), suggesting that edge-based activation functions and soft gating can better capture subtle non-linear boundaries of rare positive classes.

Second, across balanced or moderately balanced clinical datasets, KAN variants and \textsc{StandardMLP} exhibit practical equivalence. Posterior probabilities of falling within the $\pm 0.01$ ROPE interval reached unity on Breast Cancer and Chronic Kidney Disease, and exceeded $0.80$ on Heart Disease and Parkinson's Disease. Furthermore, calibration analysis showed that leading KAN variants maintain low Brier scores ($0.0580$--$0.0606$) that are on par with tuned MLPs ($0.0585$), indicating that their probability estimates are well calibrated without post-hoc recalibration.

Third, computational overhead remains a tangible trade-off for KAN architectures. On tested hardware, KAN variants incurred inference latencies that were $1.20\times$ to $1.48\times$ higher than the \textsc{StandardMLP} ($0.0611$\,ms per sample), accompanied by a $2.8\times$ to $12.5\times$ increase in trainable parameter count ($8{,}273$ parameters for the MLP vs.\ $22{,}892$--$103{,}347$ for KANs). As a result, the empirical Pareto frontier for AUPRC versus inference latency is co-defined by \textsc{StandardMLP} (minimum latency) and \textsc{TabKAN} (maximum AUPRC).

In clinical practice, these findings suggest that the choice of architecture should be guided by task requirements. For latency-sensitive applications such as emergency triage, bedside telemetry, or deployment on resource-constrained medical edge devices, \textsc{StandardMLP} remains a compelling and computationally efficient choice. Conversely, for screening tasks and clinical diagnostics where the primary objective is maximizing precision and recall on rare pathological events, architectures like \textsc{TabKAN} offer a viable and effective alternative.

Future research could explore several directions. First, given the rapid development of novel KAN architectures, expanding this comparative analysis to encompass newly emerging variants could help identify architectures specifically suited for clinical tabular data. Second, leveraging the symbolic interpretability of learned edge functions presents an opportunity for clinicians to inspect non-linear biomarker interactions directly. Finally, extending this benchmarking framework to large-scale longitudinal cohorts, such as MIMIC-IV~\cite{johnson2023mimic} and PhysioNet~\cite{goldberger2000physiobank} databases, will help assess the generalizability of these observations across diverse clinical workflows.

\section*{Declaration of generative AI and AI-assisted technologies in the manuscript preparation process}

During the preparation of this work, the authors used Gemini and Claude (Sonnet and Opus models) within the Antigravity IDE framework to assist in developing and debugging the experimental codebase (Python scripts for model evaluation). After using these tools, the authors thoroughly reviewed and edited the output as needed, verified all empirical results against the codebase, and take full responsibility for the content and integrity of the published code.

\section*{Acknowledgment}
The publication was created within the project of the Minister of Science and Higher Education "Support for the activity of Centers of Excellence established in Poland under Horizon 2020" on the basis of the contract number MEiN/2023/DIR/3796. This project has received funding from the European Union’s Horizon 2020 research and innovation programme under grant agreement No 857533.

\bibliographystyle{elsarticle-num}
\bibliography{references}

\FloatBarrier

\appendix
\clearpage
% ====================================================================
% APPENDIX A: Exhaustive Performance Benchmarks
% ====================================================================
\section{Exhaustive Performance Benchmarks}
\label{app:comprehensive_benchmark}

Here, we provide a consolidated overview of predictive performance across all seven medical tabular datasets. Table~\ref{tab:grand_summary_auprc} presents the Grand Summary across six key evaluation metrics, aggregated over $N=105$ cross-validation folds per architecture ($15$ folds across seven datasets). In addition to the primary AUPRC metric, the summary includes AUROC, Matthews Correlation Coefficient (MCC), macro-averaged F1-Score, Balanced Accuracy, and the Brier Score.

Figure~\ref{fig:app_metric_distributions} illustrates the distributional profiles across all eleven evaluated architectures for key supplementary metrics: AUROC, Matthews Correlation Coefficient (MCC), macro-averaged F1-Score, and per-sample inference latency.

% ====================================================================
% APPENDIX B: Computational Profiling
% ====================================================================
\section{Computational Profiling}
\label{app:computational_bottlenecks}

\begin{figure*}[p]
    \centering
    
    \begin{subfigure}[b]{0.58\textwidth}
        \centering
        \includegraphics[width=\textwidth]{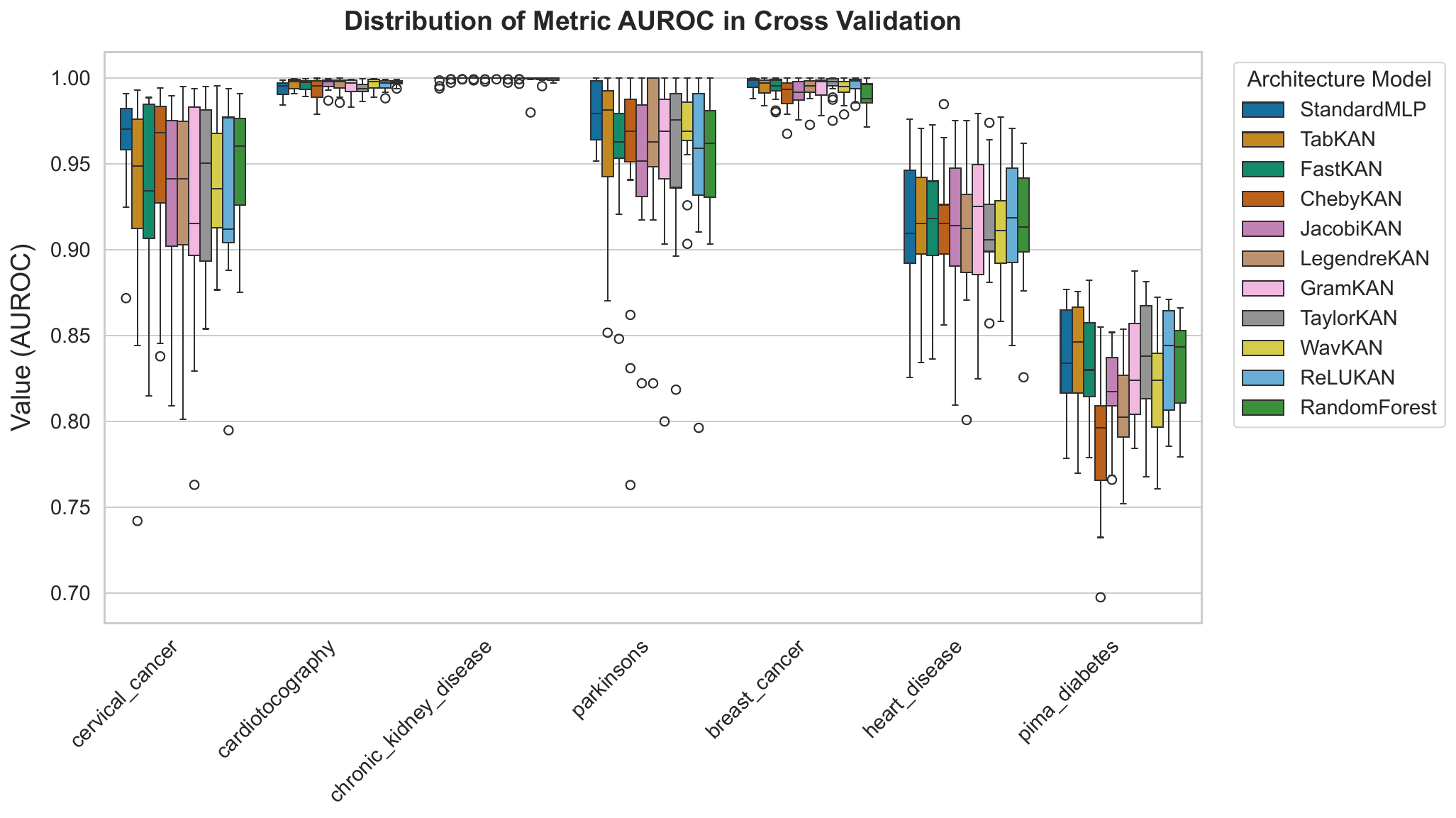}
    \end{subfigure}
    
    \vspace{0.5em}
    
    \begin{subfigure}[b]{0.58\textwidth}
        \centering
        \includegraphics[width=\textwidth]{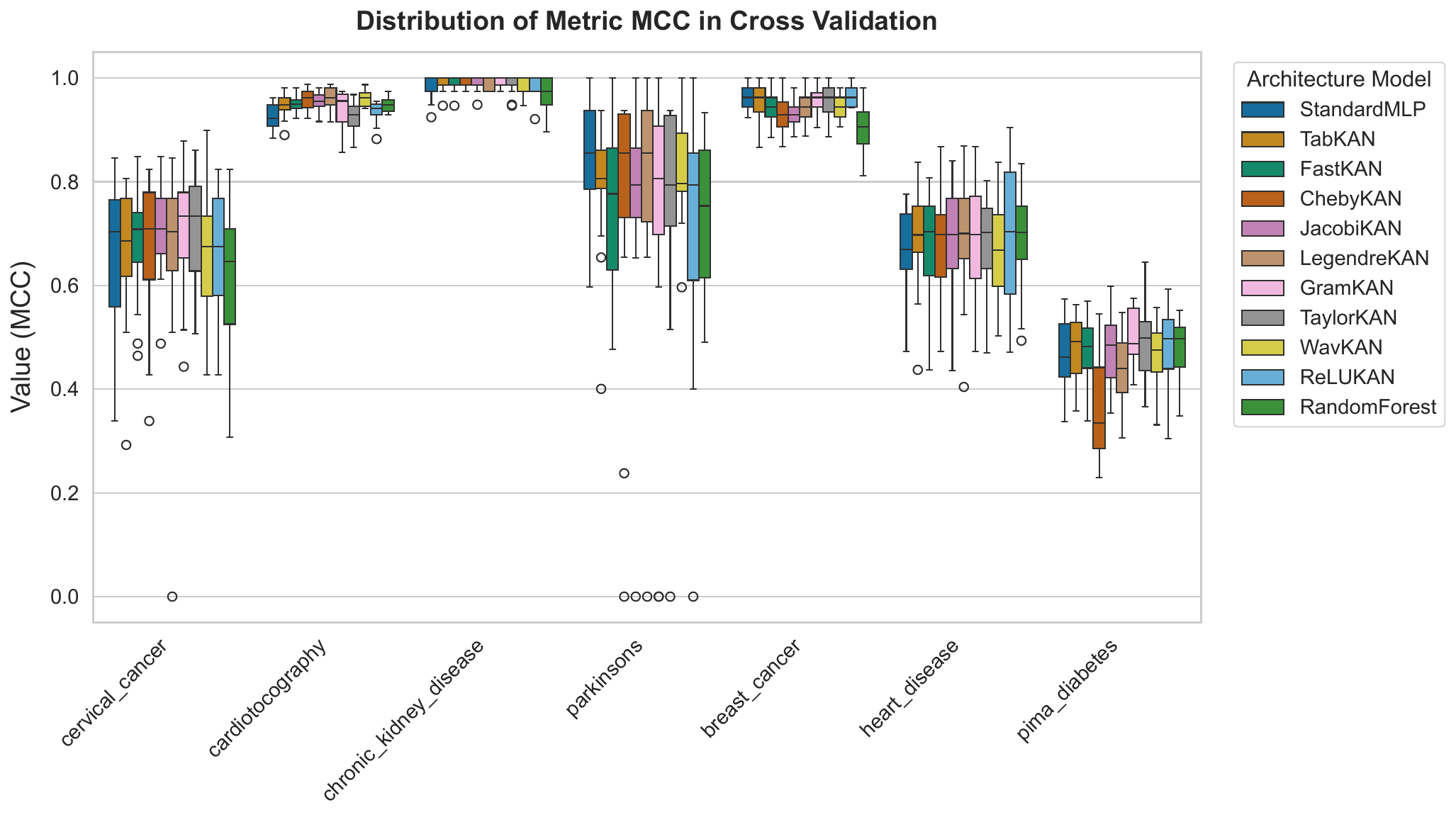}
    \end{subfigure}
    
    \vspace{0.5em}
    
    \begin{subfigure}[b]{0.58\textwidth}
        \centering
        \includegraphics[width=\textwidth]{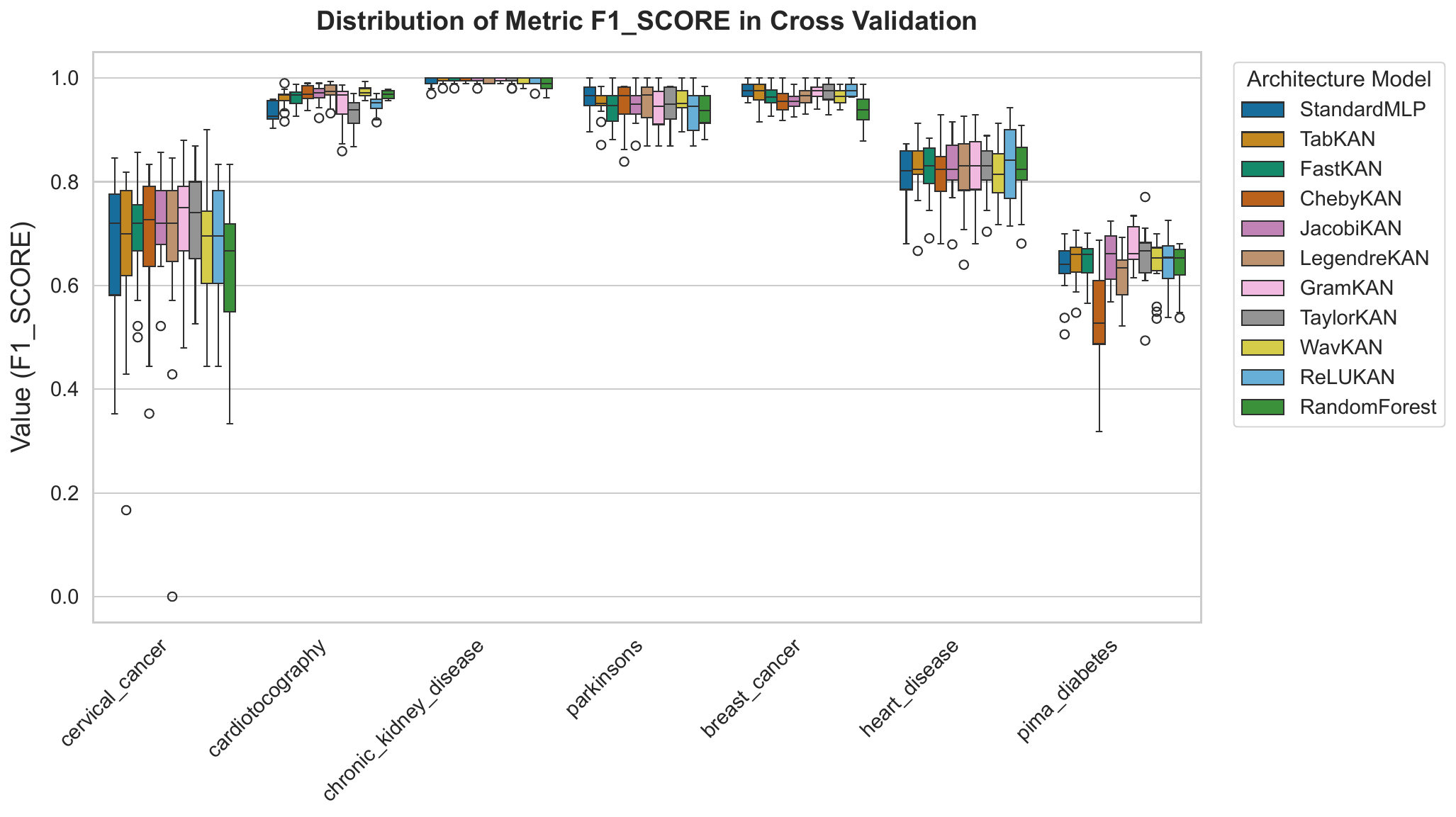}
    \end{subfigure}
    
    \vspace{0.5em}
    
    \begin{subfigure}[b]{0.58\textwidth}
        \centering
        \includegraphics[width=\textwidth]{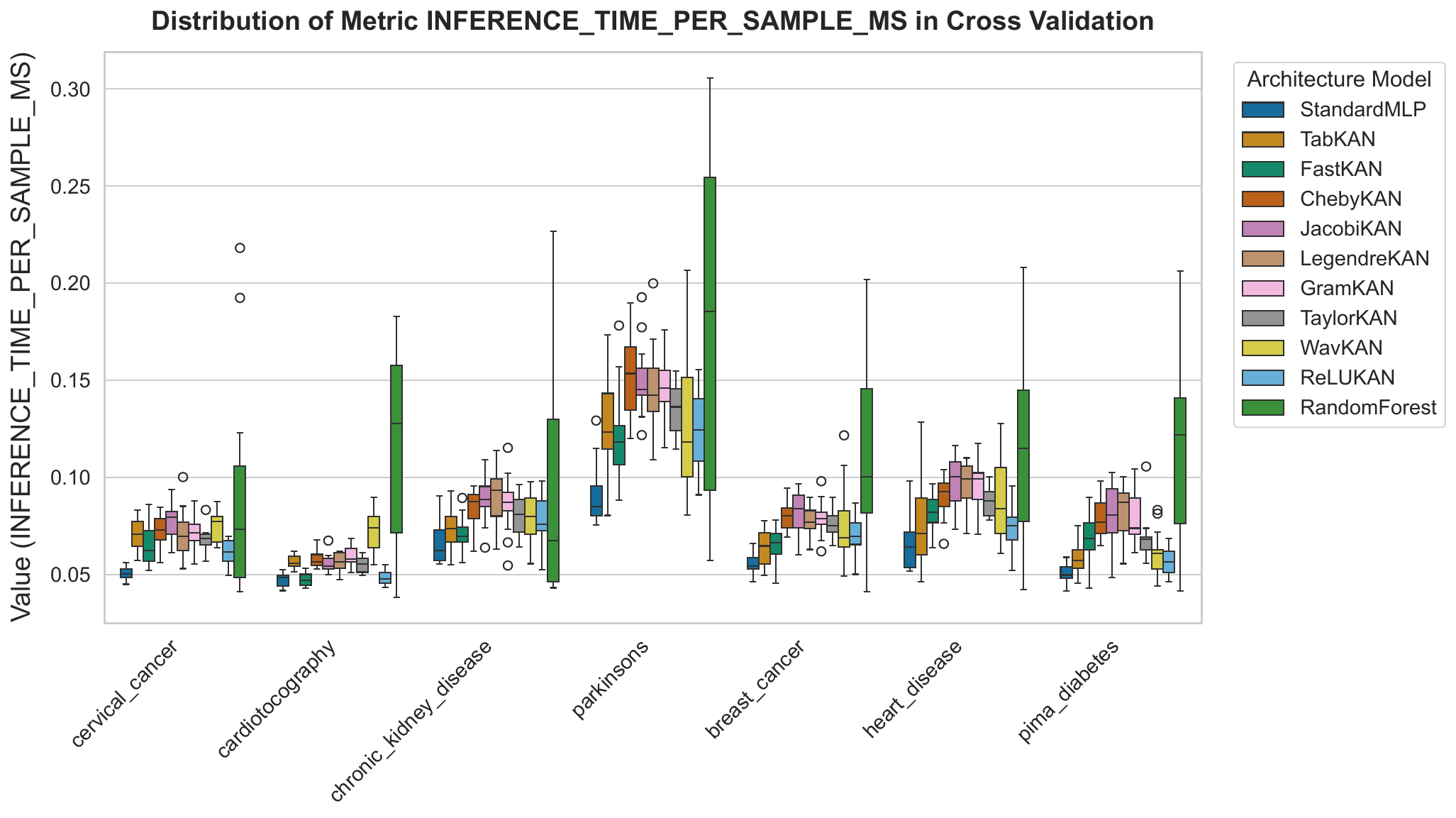}
    \end{subfigure}
    
    \caption{Gallery of cross-validation metric distributions for all evaluated architectures across seven medical datasets (from top to bottom): AUROC, MCC, macro-averaged F1-Score, and per-sample inference latency measured on Apple Silicon MPS hardware (log scale).}
    \label{fig:app_metric_distributions}
\end{figure*}

Detailed profiling of computational overhead across all architectures is summarized in Table~\ref{tab:computational_complexity}. Figure~\ref{fig:app_metric_distributions} (bottom panel) presents the per-sample inference latency distributions on a logarithmic scale.

The empirical latency disparity is attributable to computational primitives: \textsc{StandardMLP} computes layer outputs using heavily optimized General Matrix Multiply (GEMM) kernels with scalar weights, which fully leverage the hardware architecture of modern processors. In contrast, KAN architectures evaluate univariate basis functions along network edges, requiring additional memory bandwidth and kernel dispatches. While specialized variants like \textsc{ReLUKAN} and \textsc{TabKAN} mitigate this gap compared to higher-degree polynomial KANs, the structural latency overhead remains an important consideration for real-time clinical deployment.

% ====================================================================
% APPENDIX C: Sample Efficiency and Data Regime Ablation
% ====================================================================
\section{Sample Efficiency and Data Regime Ablation}
\label{app:ablation}

\begin{figure*}[p]
    \centering
    
    % --- Row 1 ---
    \begin{subfigure}[b]{0.48\textwidth}
        \centering
        \includegraphics[width=\textwidth]{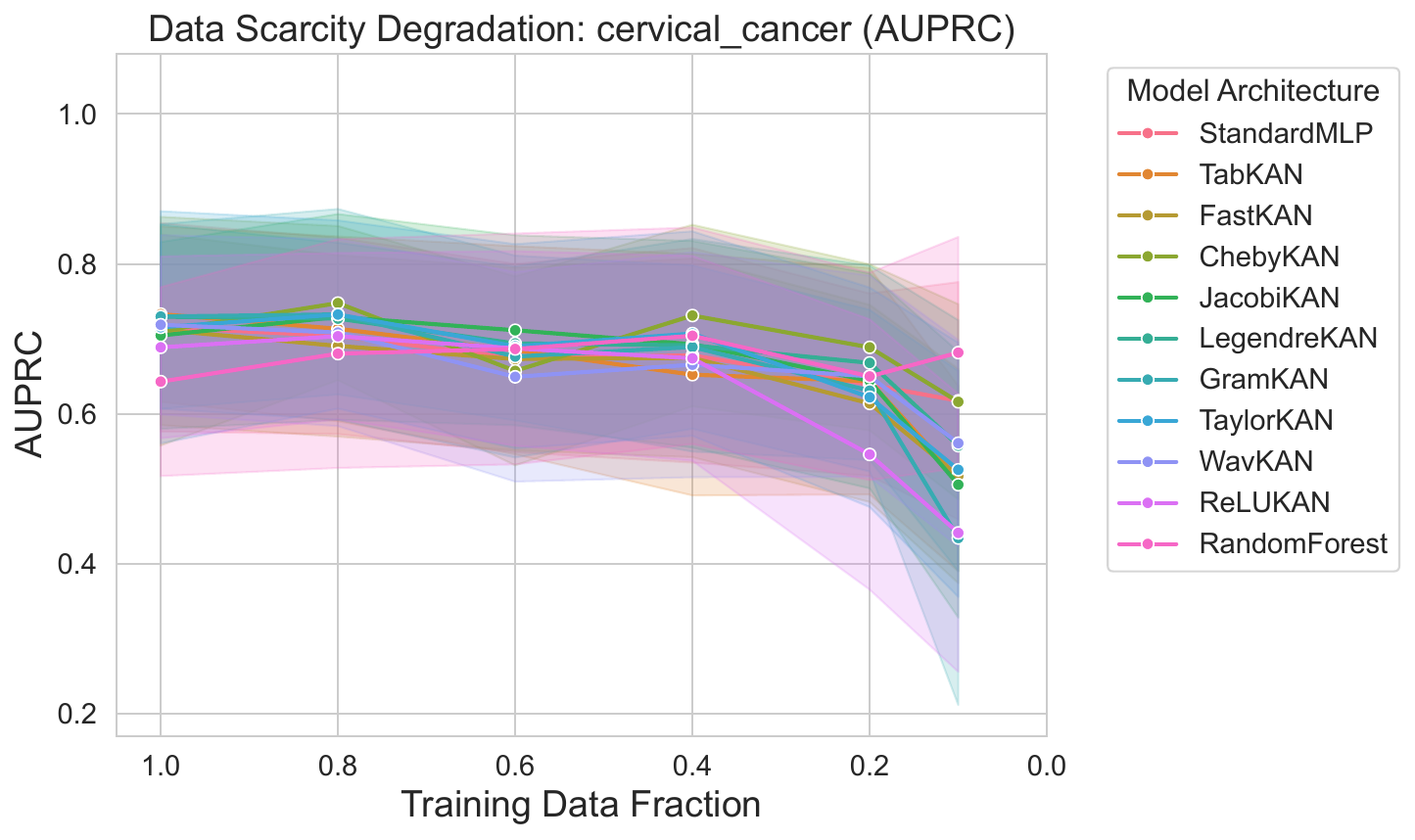}
    \end{subfigure}
    \hfill
    \begin{subfigure}[b]{0.48\textwidth}
        \centering
        \includegraphics[width=\textwidth]{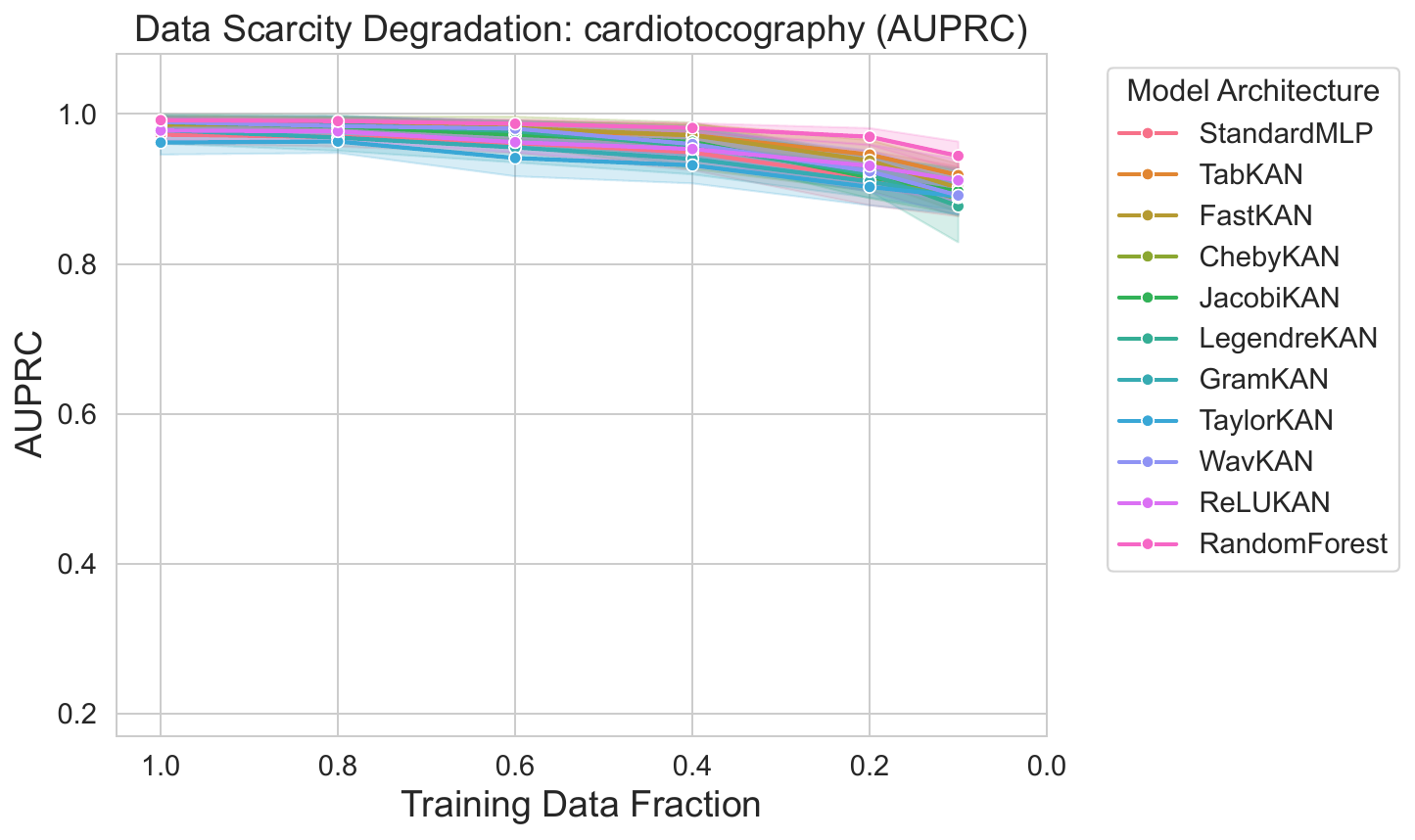}
    \end{subfigure}
    
    \vspace{2em}
    
    % --- Row 2 ---
    \begin{subfigure}[b]{0.48\textwidth}
        \centering
        \includegraphics[width=\textwidth]{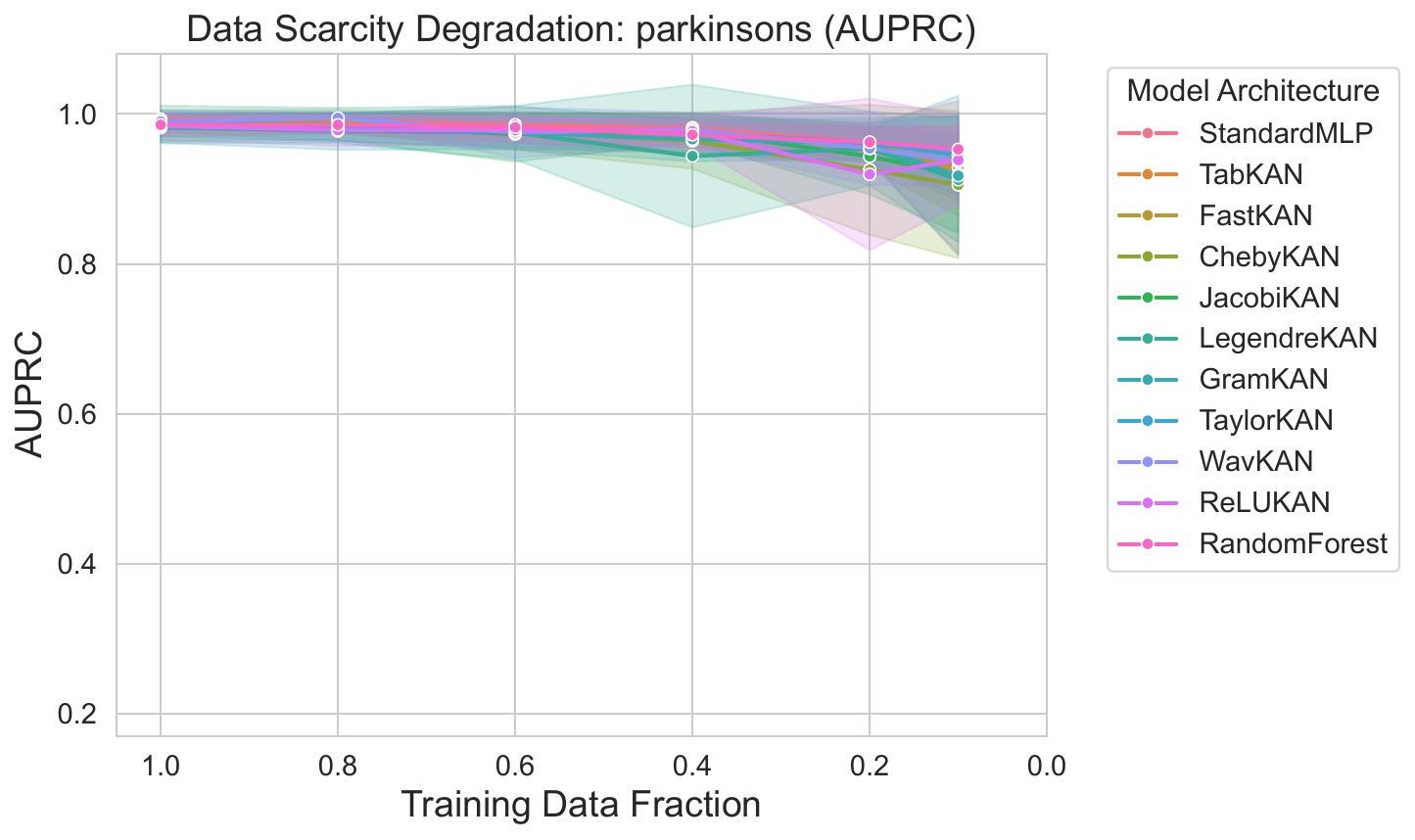}
    \end{subfigure}
    \hfill
    \begin{subfigure}[b]{0.48\textwidth}
        \centering
        \includegraphics[width=\textwidth]{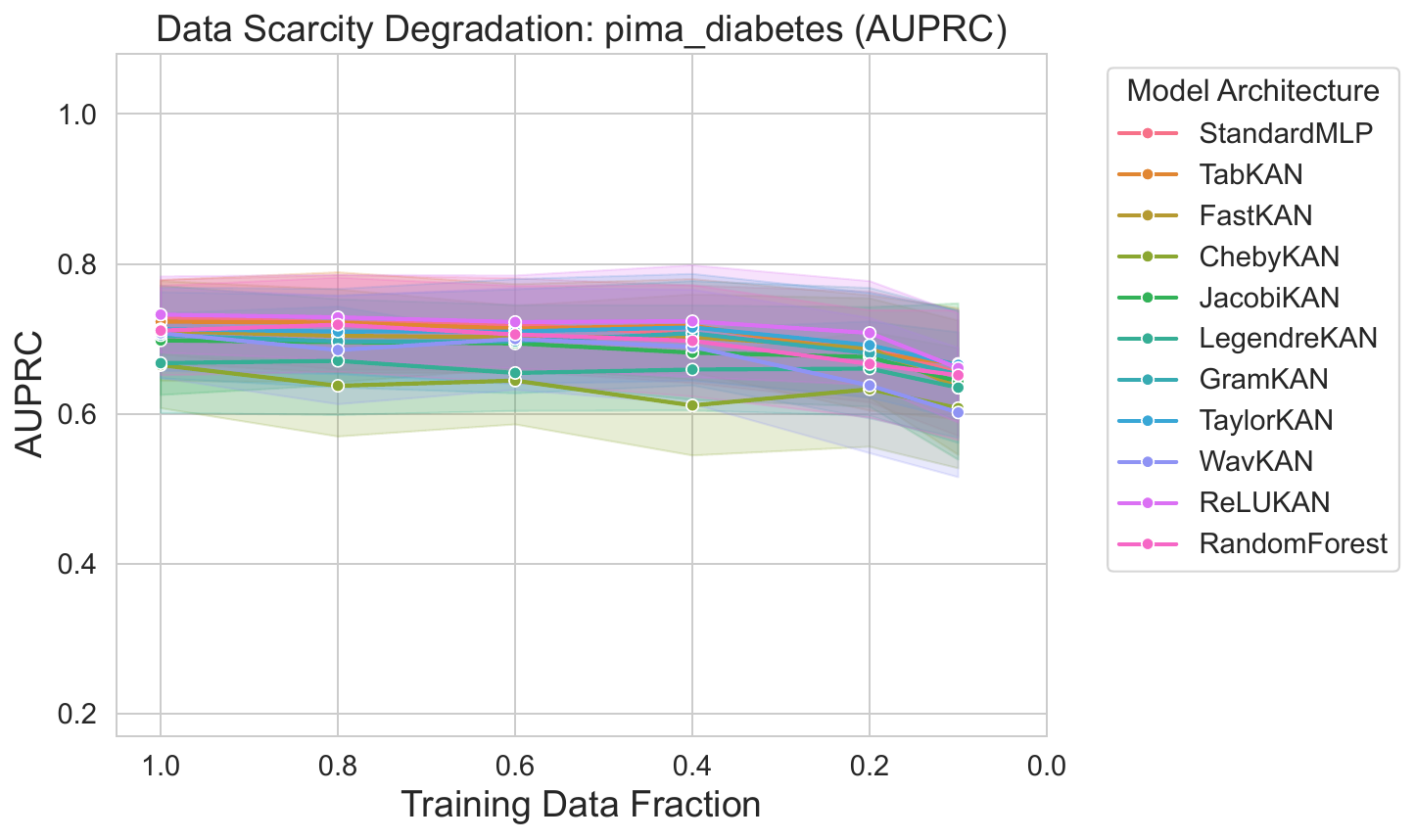}
    \end{subfigure}
    
    \vspace{2em}
    
    % --- Row 3 ---
    \begin{subfigure}[b]{0.48\textwidth}
        \centering
        \includegraphics[width=\textwidth]{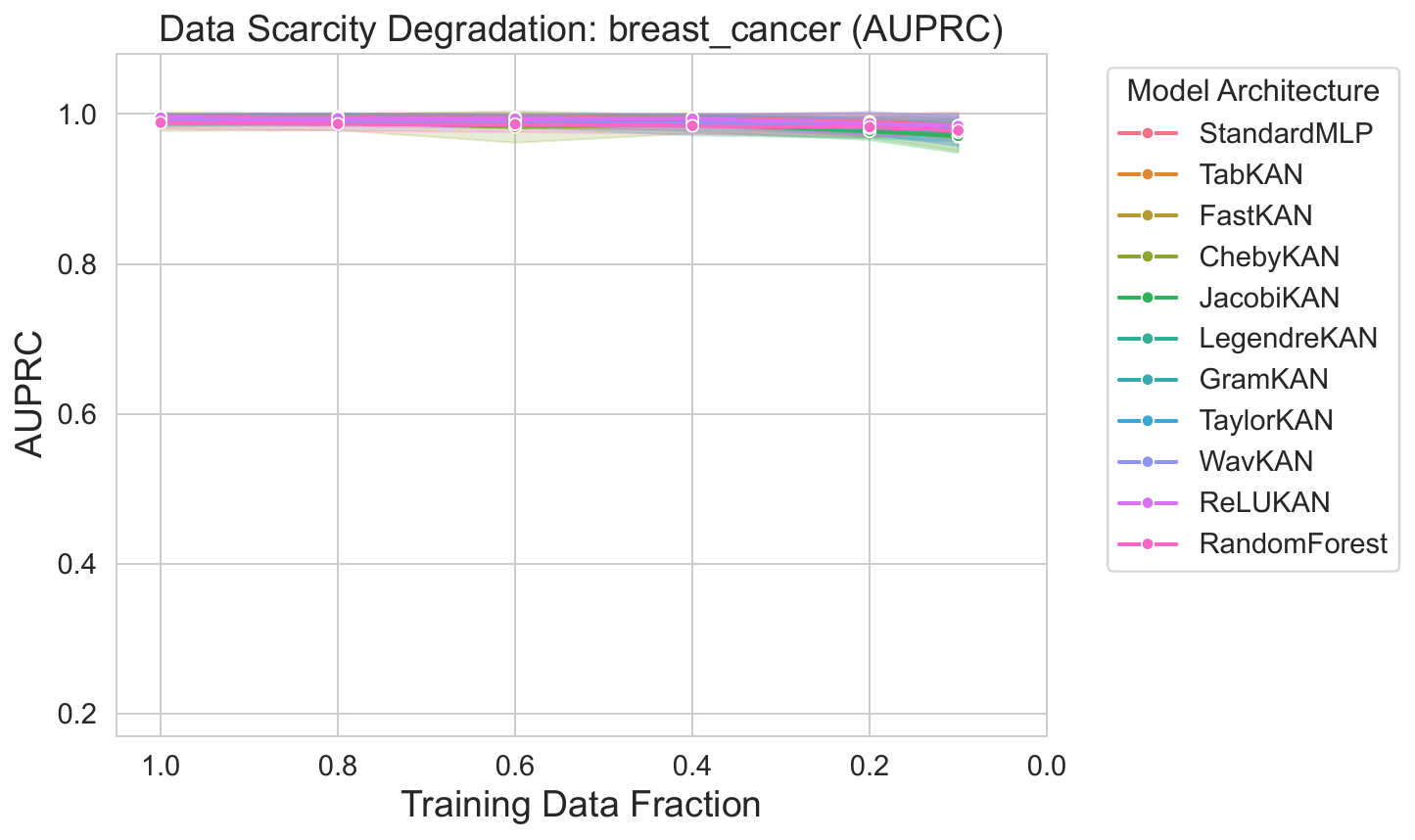}
    \end{subfigure}
    \hfill
    \begin{subfigure}[b]{0.48\textwidth}
        \centering
        \includegraphics[width=\textwidth]{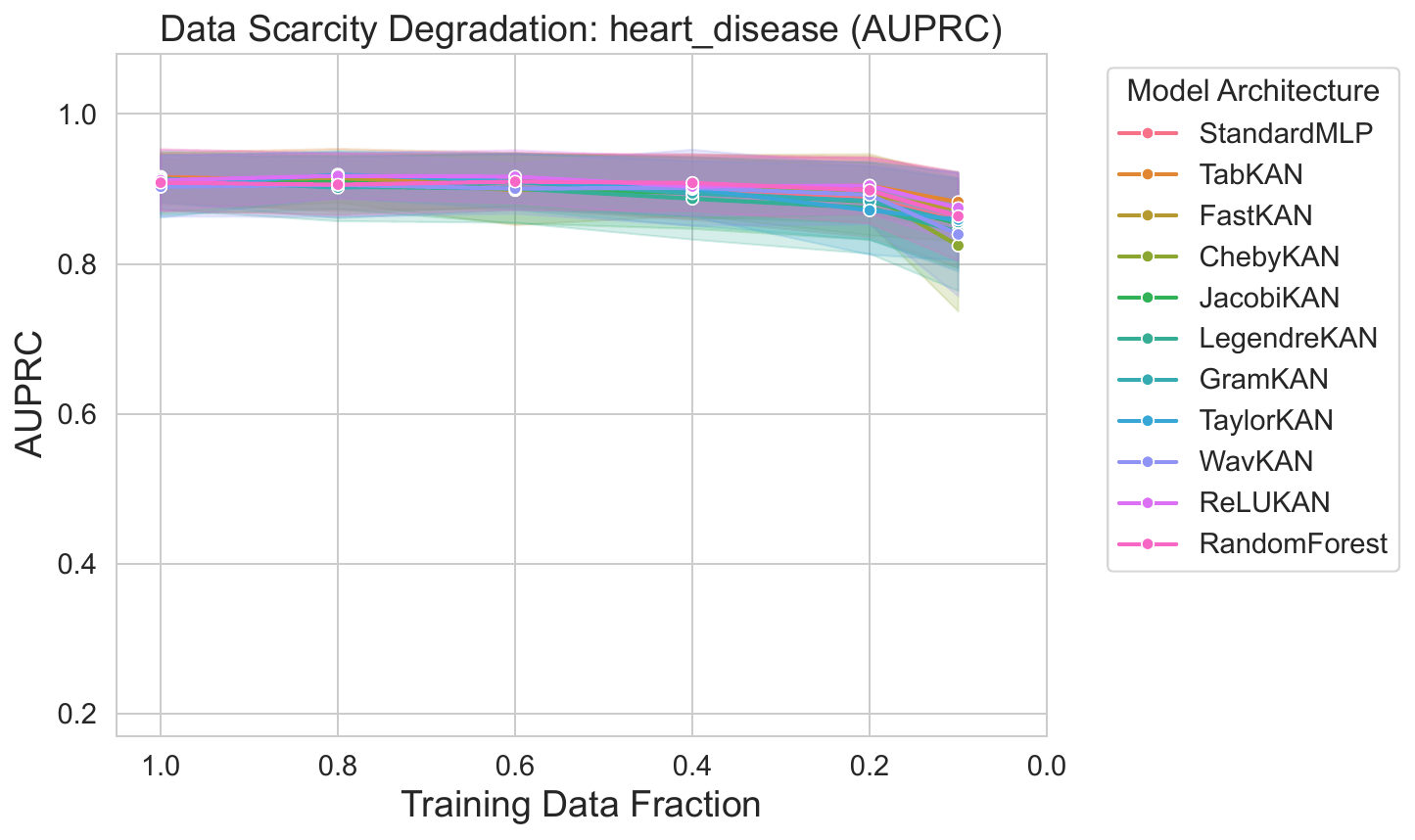}
    \end{subfigure}
    
    \caption{Sample efficiency ablation curves illustrating AUPRC degradation as training data is reduced from $100\%$ to $10\%$ across six clinical cohorts (left to right, top to bottom): Cervical Cancer, Cardiotocography, Parkinson's Disease, Pima Diabetes, Breast Cancer, and Heart Disease. Shaded bands denote $\pm 1$ standard deviation across folds.}
    \label{fig:app_ablation}
\end{figure*}

To examine how architectural choices influence resilience to data scarcity, we evaluated model performance in terms of AUPRC as the training sample size was progressively ablated from $100\%$ down to $10\%$ of available training data. Figure~\ref{fig:app_ablation} displays the degradation trajectories across six representative datasets spanning high, moderate, and low class imbalance: Cervical Cancer, Cardiotocography, Parkinson's Disease, Pima Diabetes, Breast Cancer, and Heart Disease.

Across all evaluated models, performance decreases steadily as the training data fraction is reduced. All tested models exhibit comparable degradation slopes, suggesting that the non-linear basis expansions of KANs neither accelerate sample efficiency in severely restricted regimes nor introduce excessive sample inefficiency relative to a well-regularized MLP.

% ====================================================================
% APPENDIX D: Training Dynamics and Convergence Stability
% ====================================================================
\section{Training Dynamics and Convergence Stability}
\label{app:training_dynamics}

\begin{figure*}[htbp]
    \centering
    % Row 1: Cervical Cancer
    \begin{subfigure}[b]{0.48\textwidth}
        \centering
        \includegraphics[width=\textwidth]{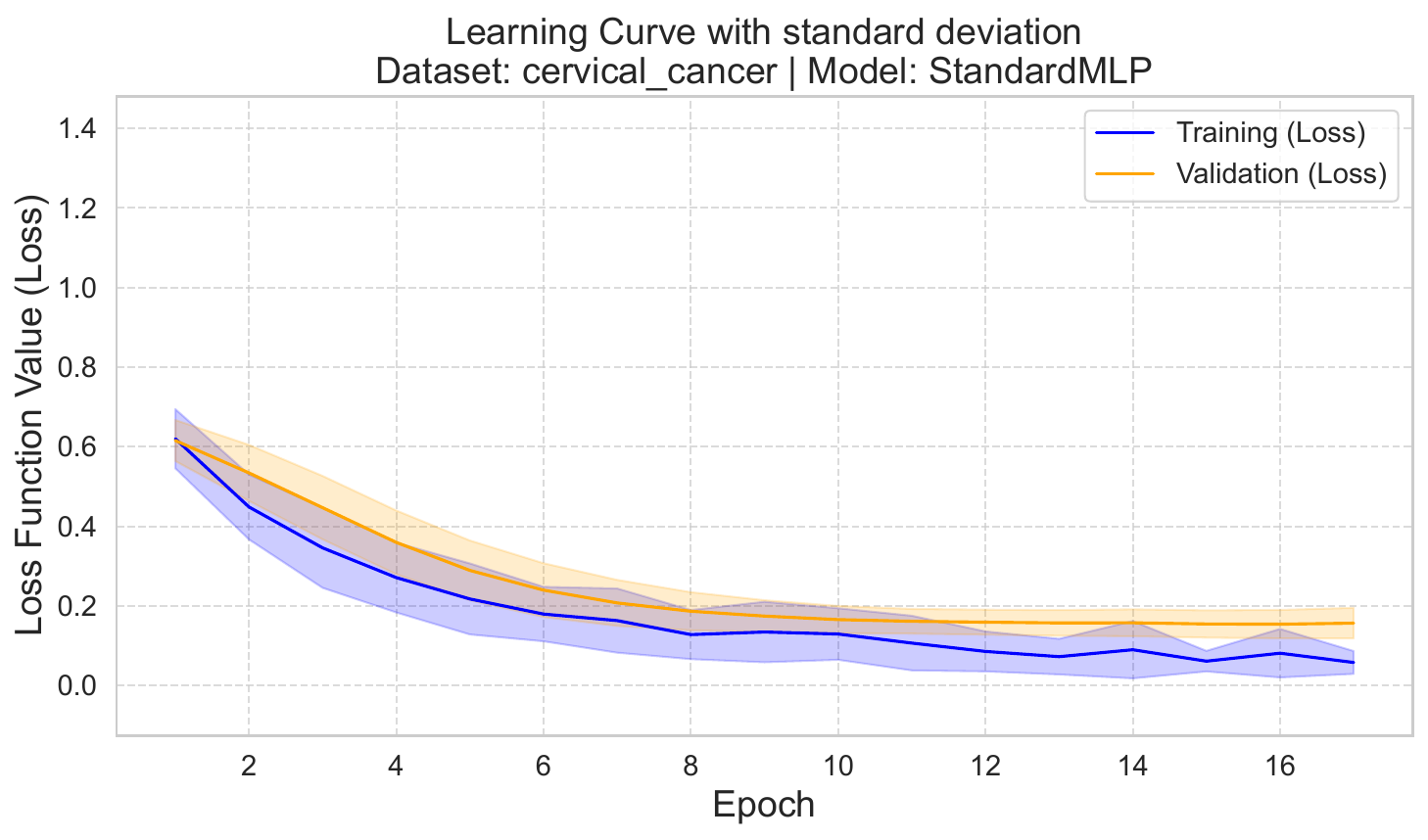}
    \end{subfigure}
    \hfill
    \begin{subfigure}[b]{0.48\textwidth}
        \centering
        \includegraphics[width=\textwidth]{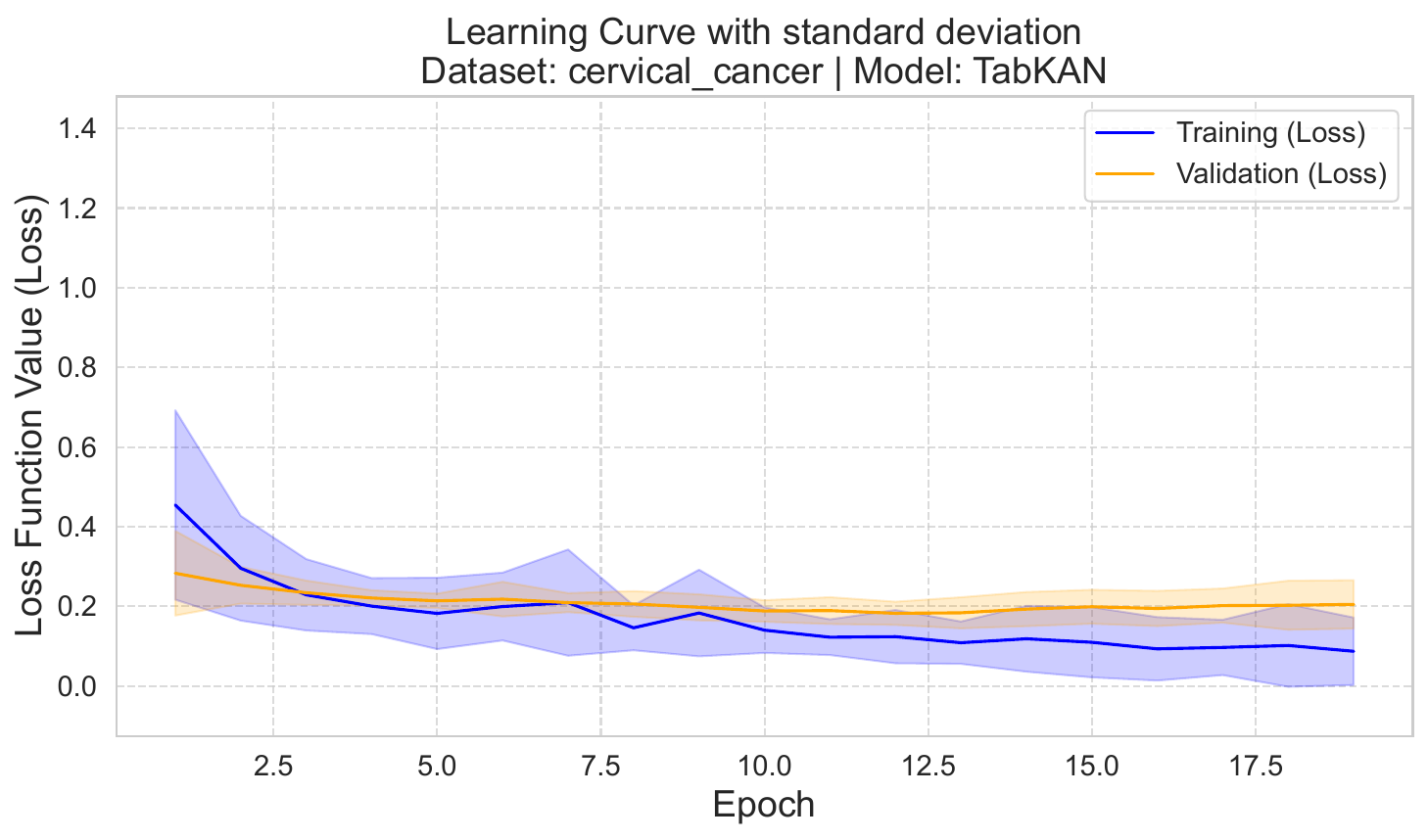}
    \end{subfigure}
    
    \vspace{3mm}
    % Row 2: Cardiotocography
    \begin{subfigure}[b]{0.48\textwidth}
        \centering
        \includegraphics[width=\textwidth]{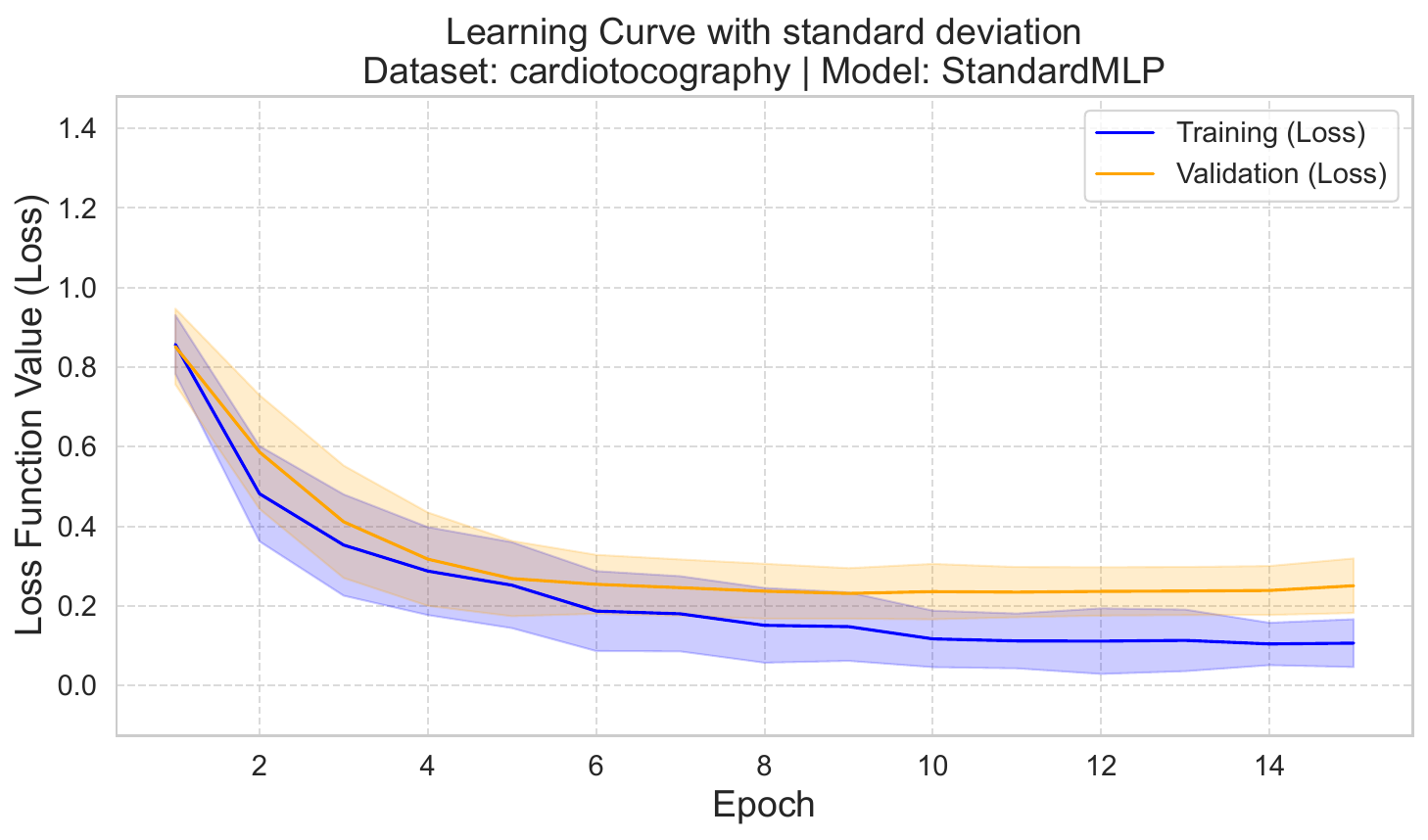}
    \end{subfigure}
    \hfill
    \begin{subfigure}[b]{0.48\textwidth}
        \centering
        \includegraphics[width=\textwidth]{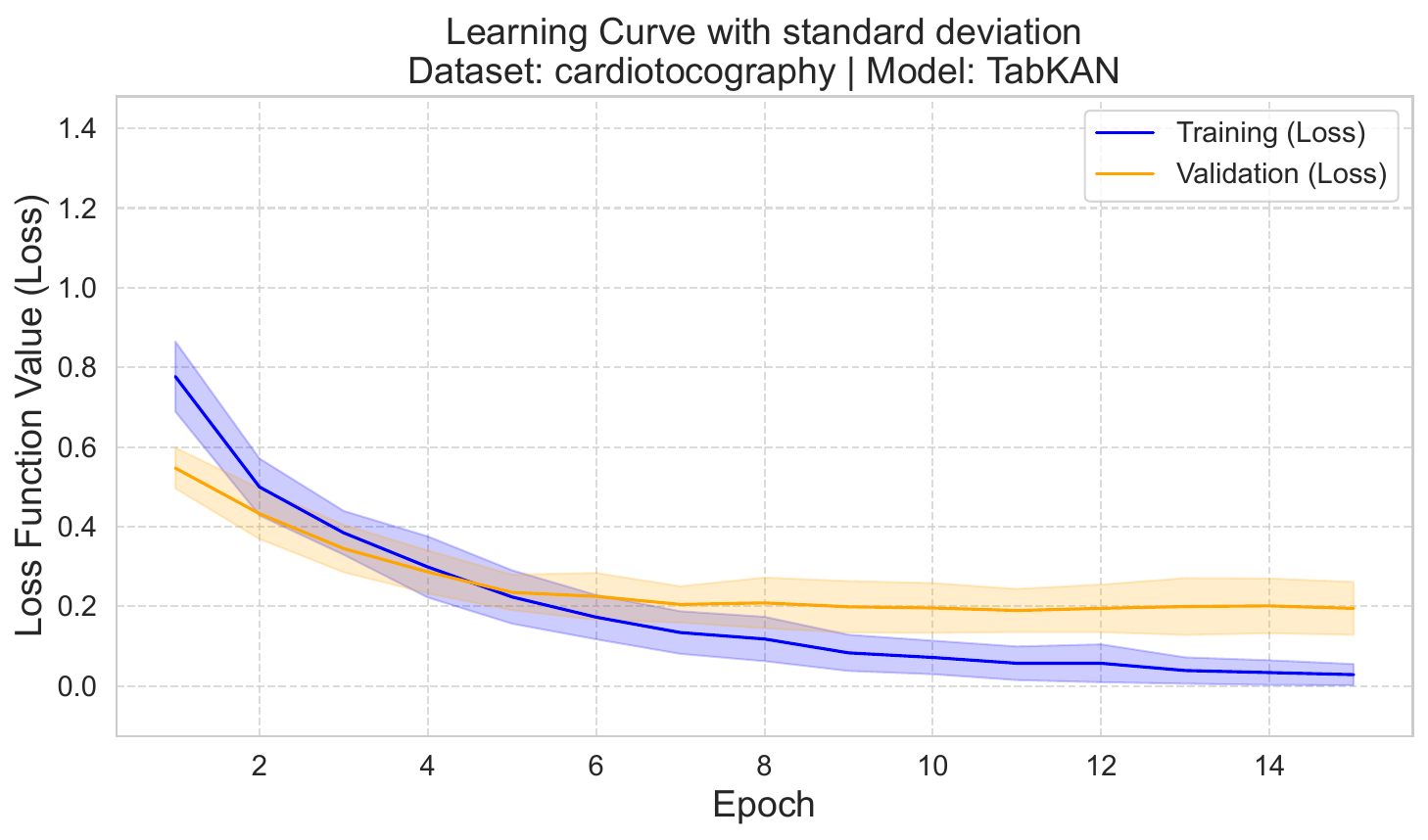}
    \end{subfigure}
    
    \vspace{3mm}
    % Row 3: Heart Disease
    \begin{subfigure}[b]{0.48\textwidth}
        \centering
        \includegraphics[width=\textwidth]{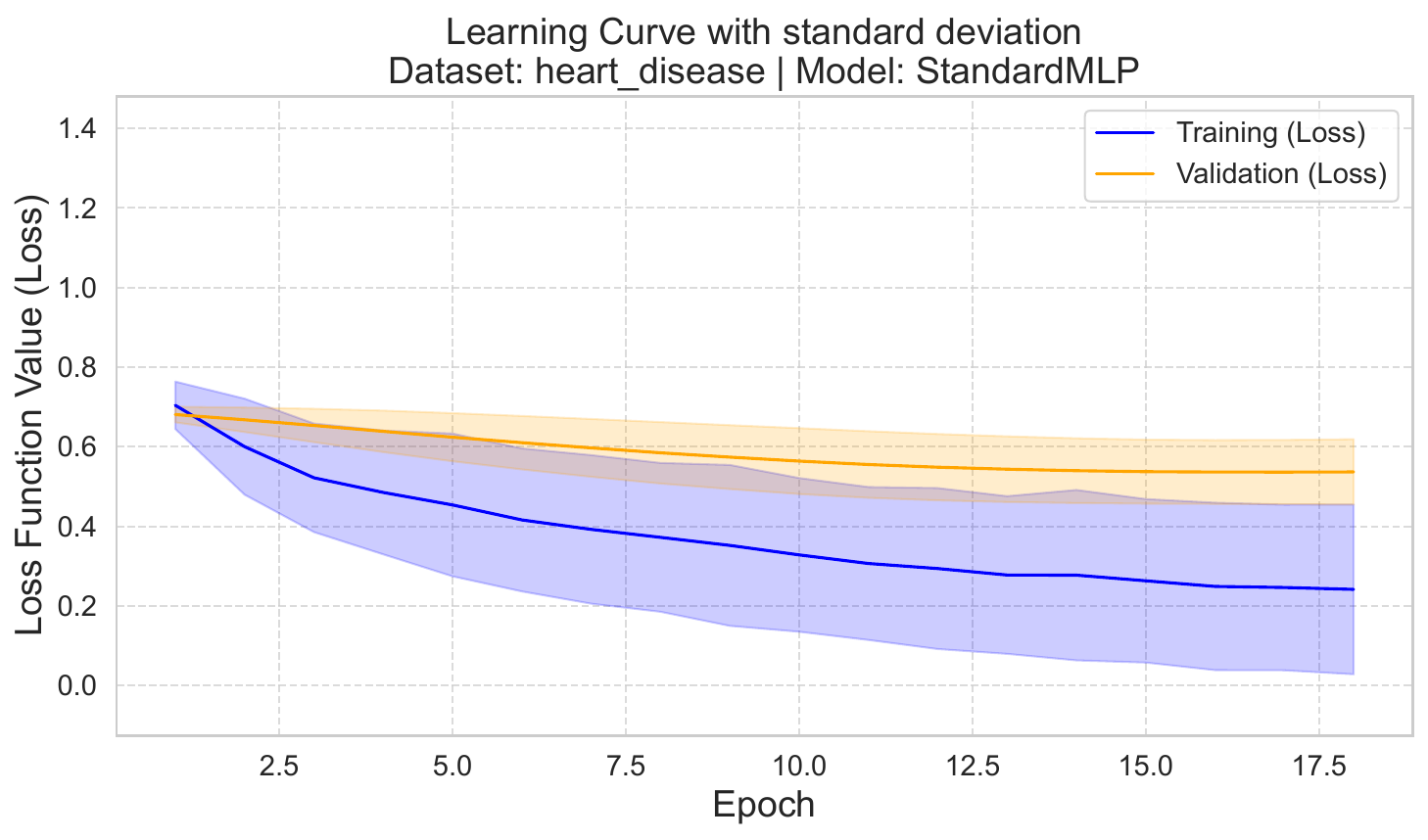}
    \end{subfigure}
    \hfill
    \begin{subfigure}[b]{0.48\textwidth}
        \centering
        \includegraphics[width=\textwidth]{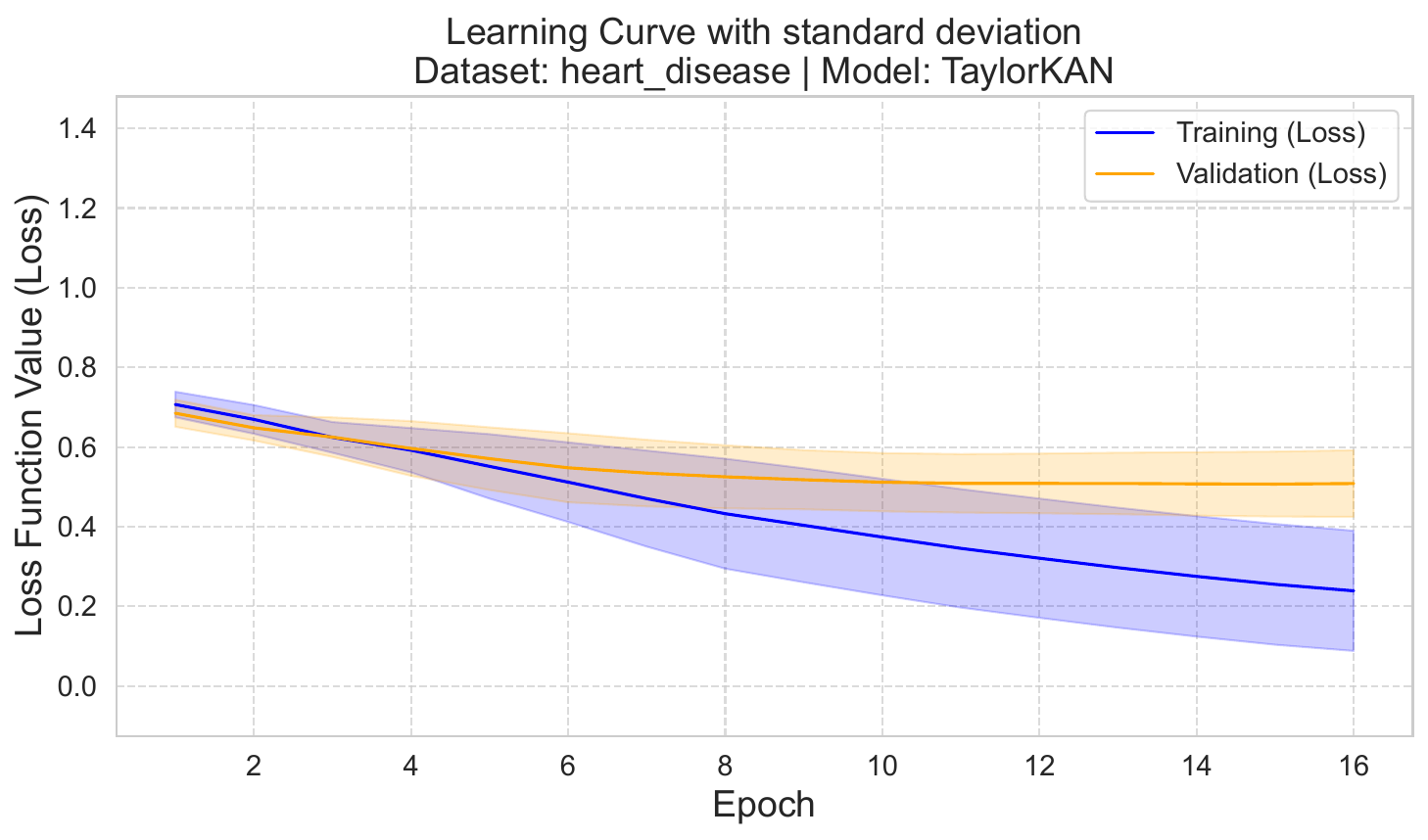}
    \end{subfigure}
    
    \vspace{3mm}
    % Row 4: Breast Cancer
    \begin{subfigure}[b]{0.48\textwidth}
        \centering
        \includegraphics[width=\textwidth]{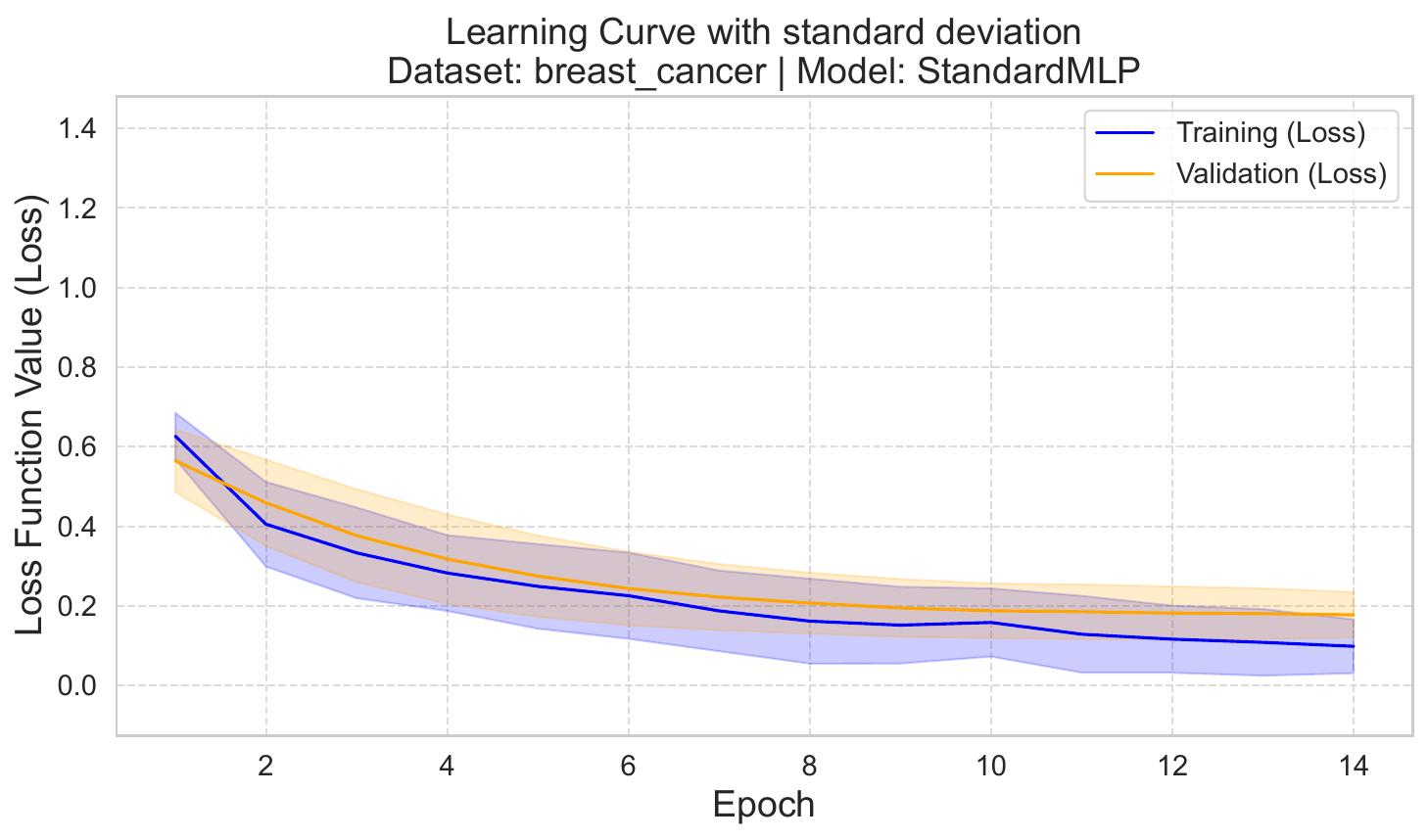}
    \end{subfigure}
    \hfill
    \begin{subfigure}[b]{0.48\textwidth}
        \centering
        \includegraphics[width=\textwidth]{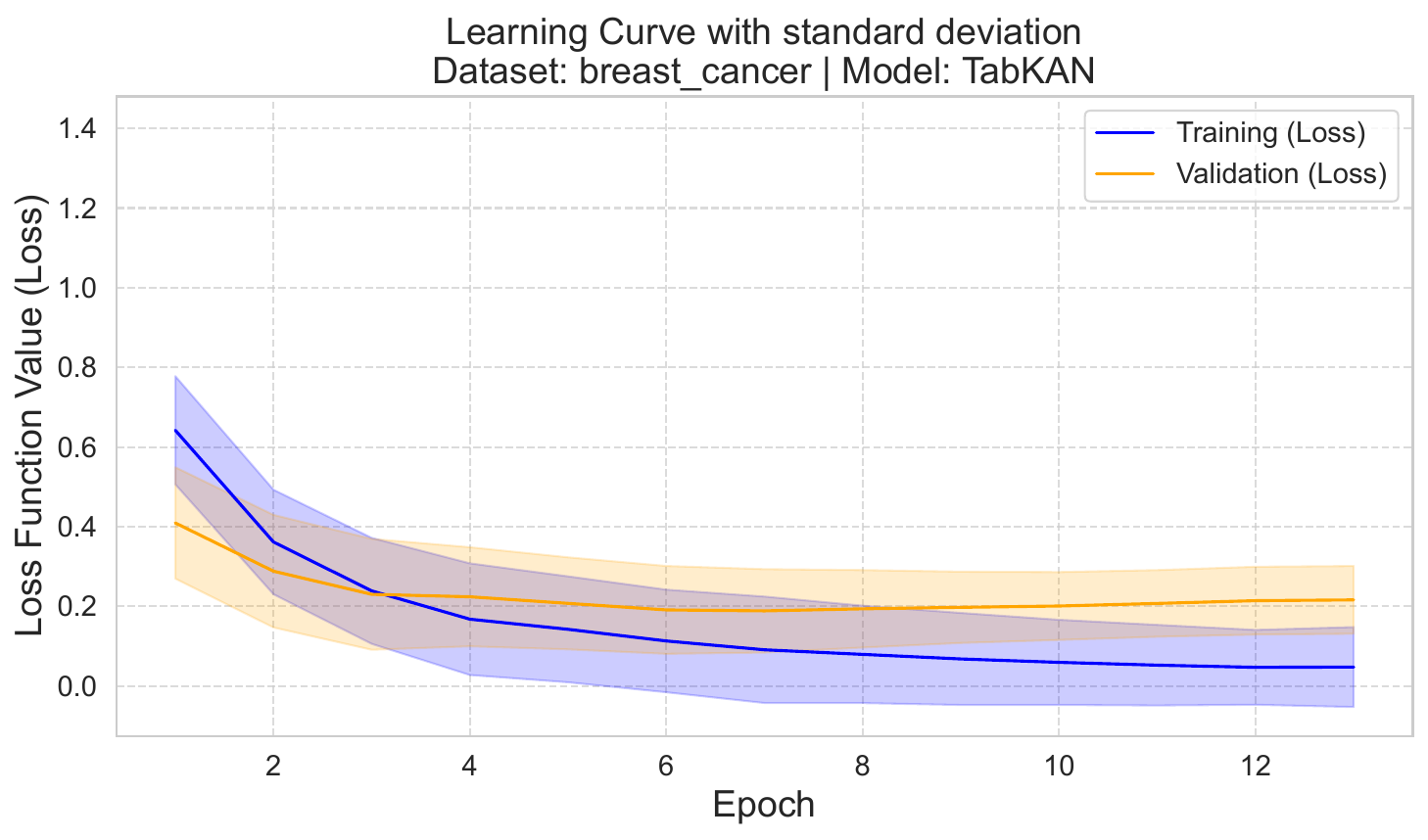}
    \end{subfigure}
    \caption{Training and validation loss trajectories across four distinct clinical datasets, comparing baseline \textsc{StandardMLP} against representative KAN variants (separate columns). Rows from top to bottom: Cervical Cancer ($6.4\%$ positive prevalence), Cardiotocography ($8.3\%$ minority class), Heart Disease, and Breast Cancer.}
    \label{fig:app_lc_grid}
\end{figure*}

Convergence behavior across training epochs was recorded to assess optimization stability. Figure~\ref{fig:app_lc_grid} presents training and validation loss curves comparing the baseline \textsc{StandardMLP} with representative KAN variants across four clinical domains.

Both model families converge reliably within the allocated epoch budget under AdamW optimization. Due to their flexible univariate activations, KAN variants often exhibit slightly steeper initial training loss reductions. However, validation loss curves plateau at similar epochs, highlighting the importance of the early-stopping mechanism implemented in our experimental protocol to prevent overparameterized KAN variants from overfitting on smaller clinical cohorts.

% ====================================================================
% APPENDIX E: Diagnostic Confusion Matrices
% ====================================================================
\section{Diagnostic Confusion Matrices}
\label{app:failure_modes}

\begin{figure*}[htbp]
    \centering
    % Row 1: Cervical Cancer
    \begin{subfigure}[b]{0.48\textwidth}
        \centering
        \includegraphics[width=0.8\textwidth]{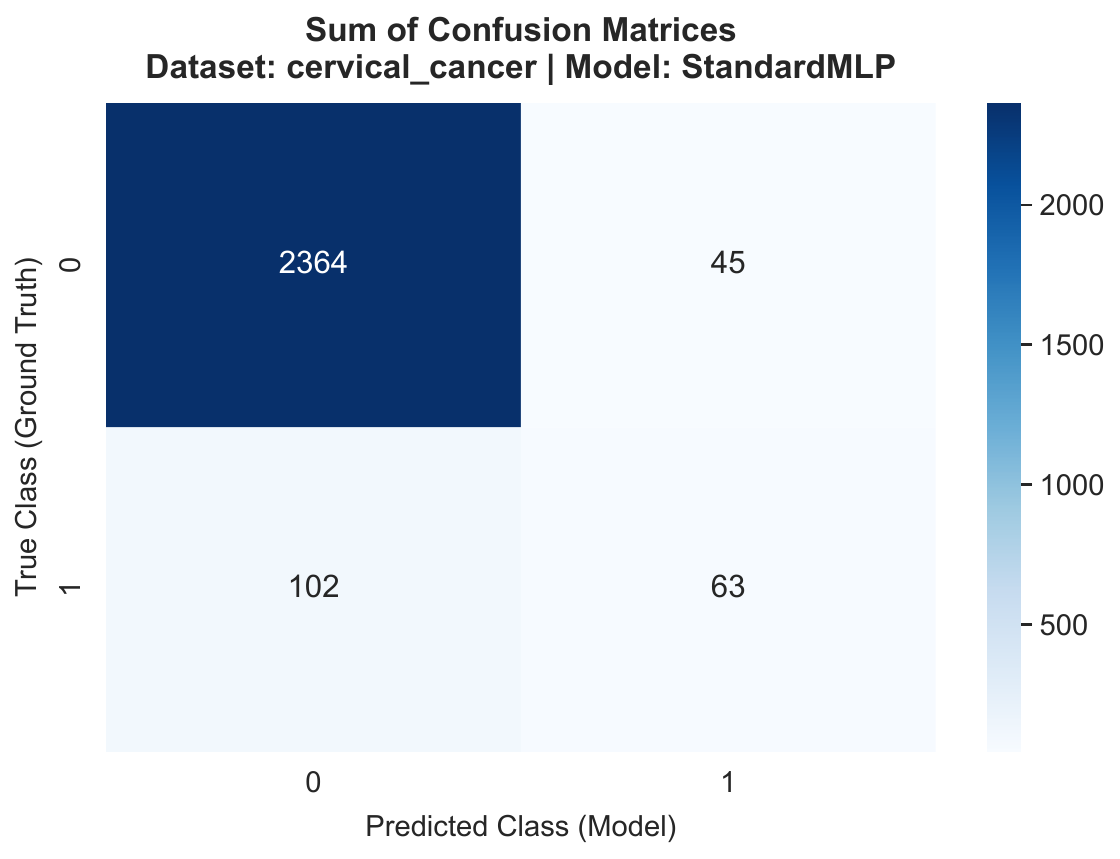}
    \end{subfigure}
    \hfill
    \begin{subfigure}[b]{0.48\textwidth}
        \centering
        \includegraphics[width=0.8\textwidth]{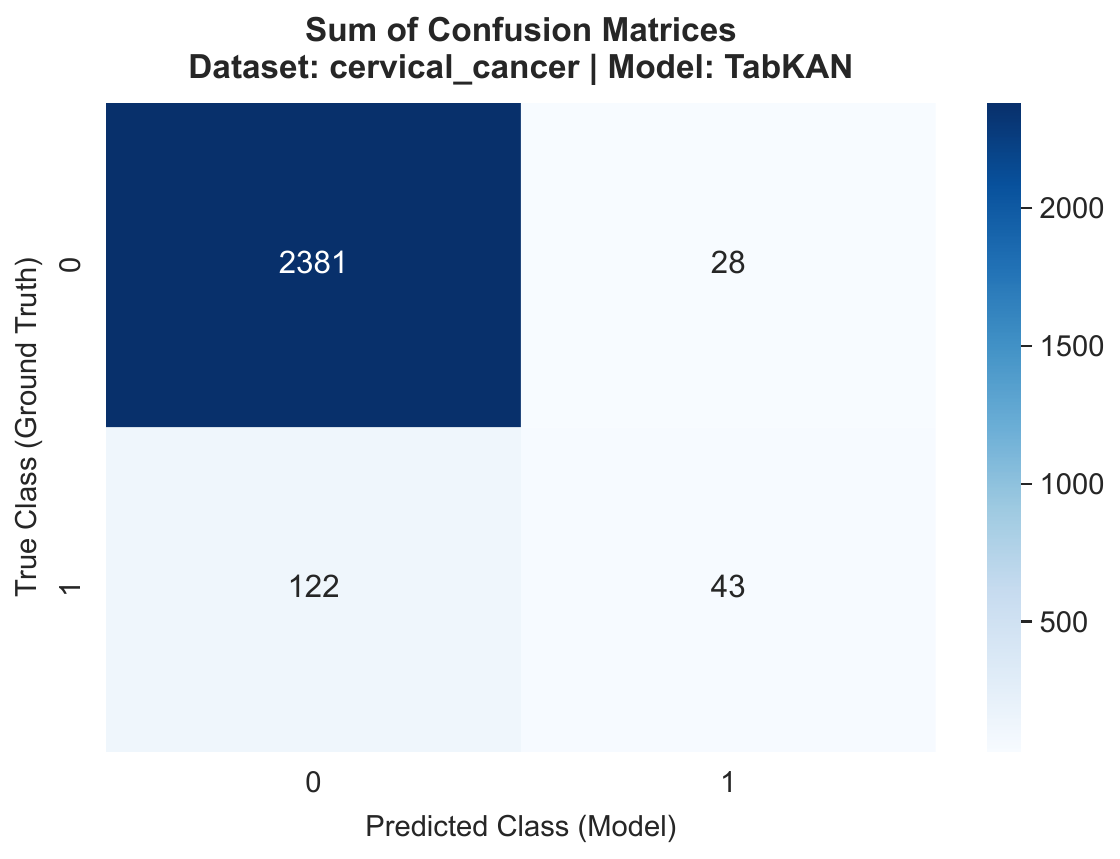}
    \end{subfigure}
    
    \vspace{3mm}
    % Row 2: Cardiotocography
    \begin{subfigure}[b]{0.48\textwidth}
        \centering
        \includegraphics[width=0.8\textwidth]{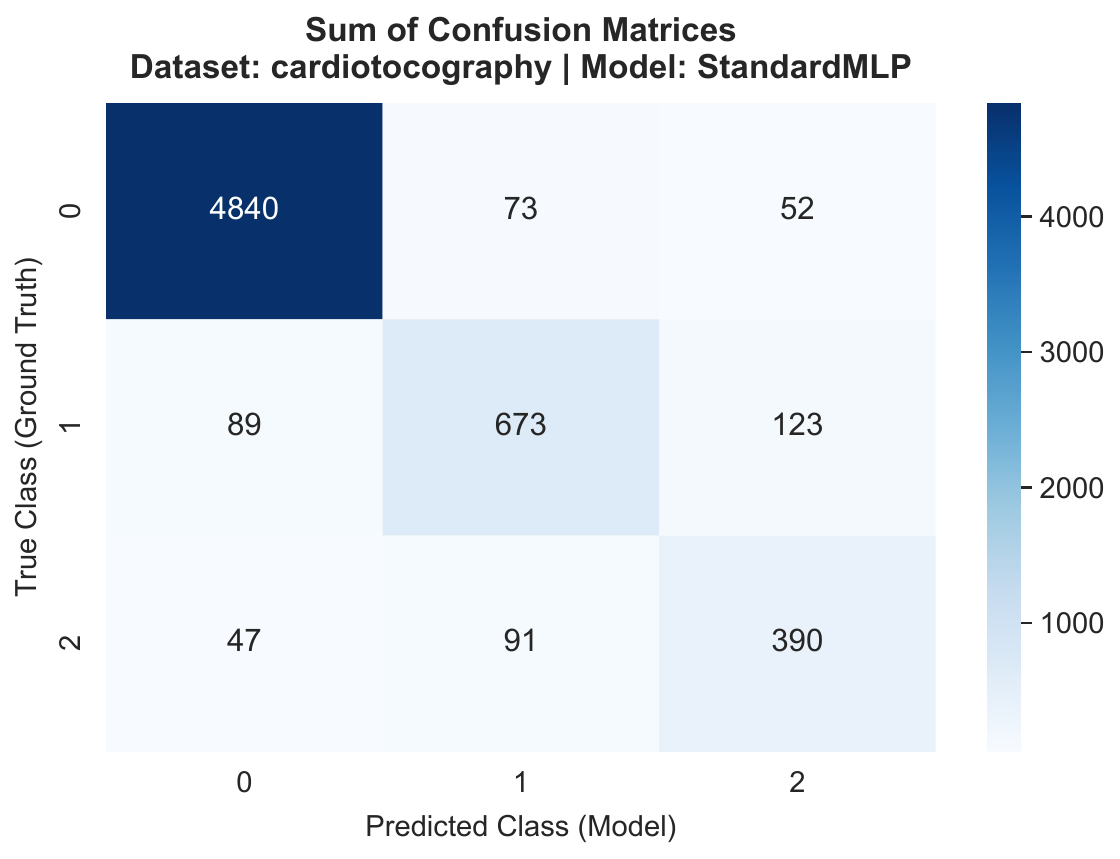}
    \end{subfigure}
    \hfill
    \begin{subfigure}[b]{0.48\textwidth}
        \centering
        \includegraphics[width=0.8\textwidth]{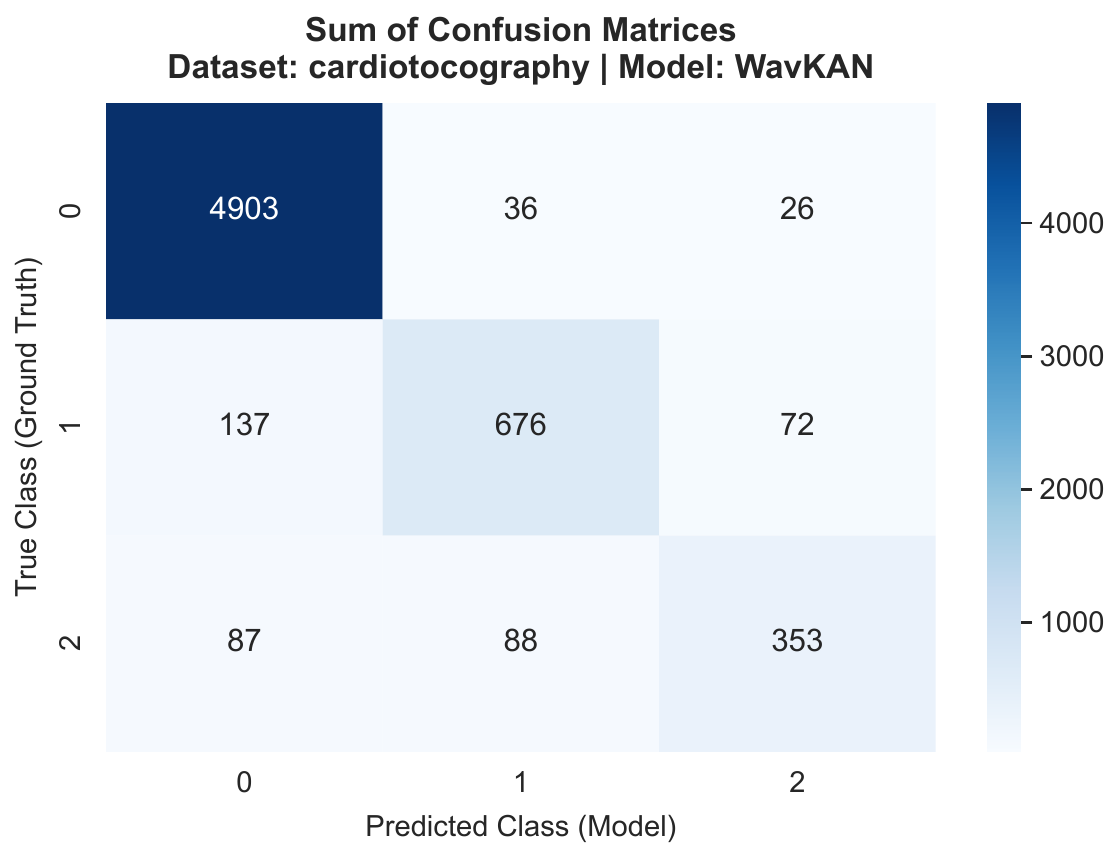}
    \end{subfigure}
    
    \vspace{3mm}
    % Row 3: Pima Diabetes
    \begin{subfigure}[b]{0.48\textwidth}
        \centering
        \includegraphics[width=0.8\textwidth]{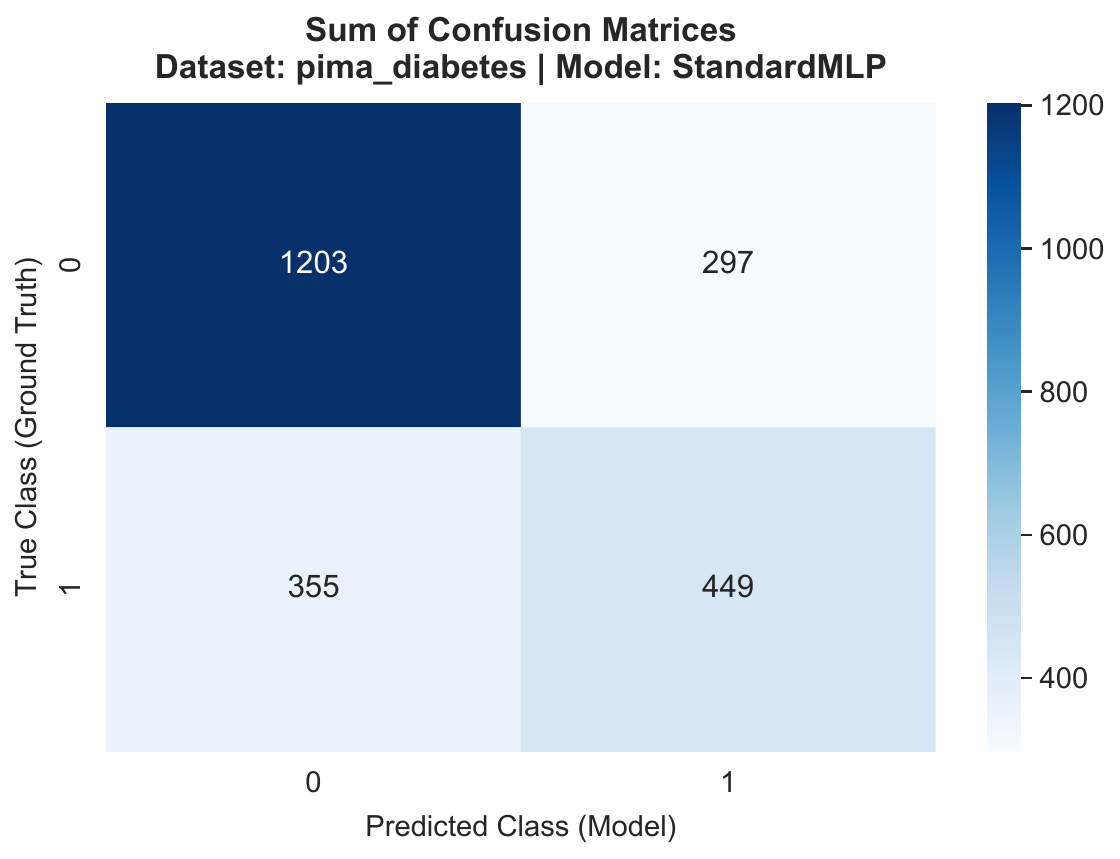}
    \end{subfigure}
    \hfill
    \begin{subfigure}[b]{0.48\textwidth}
        \centering
        \includegraphics[width=0.8\textwidth]{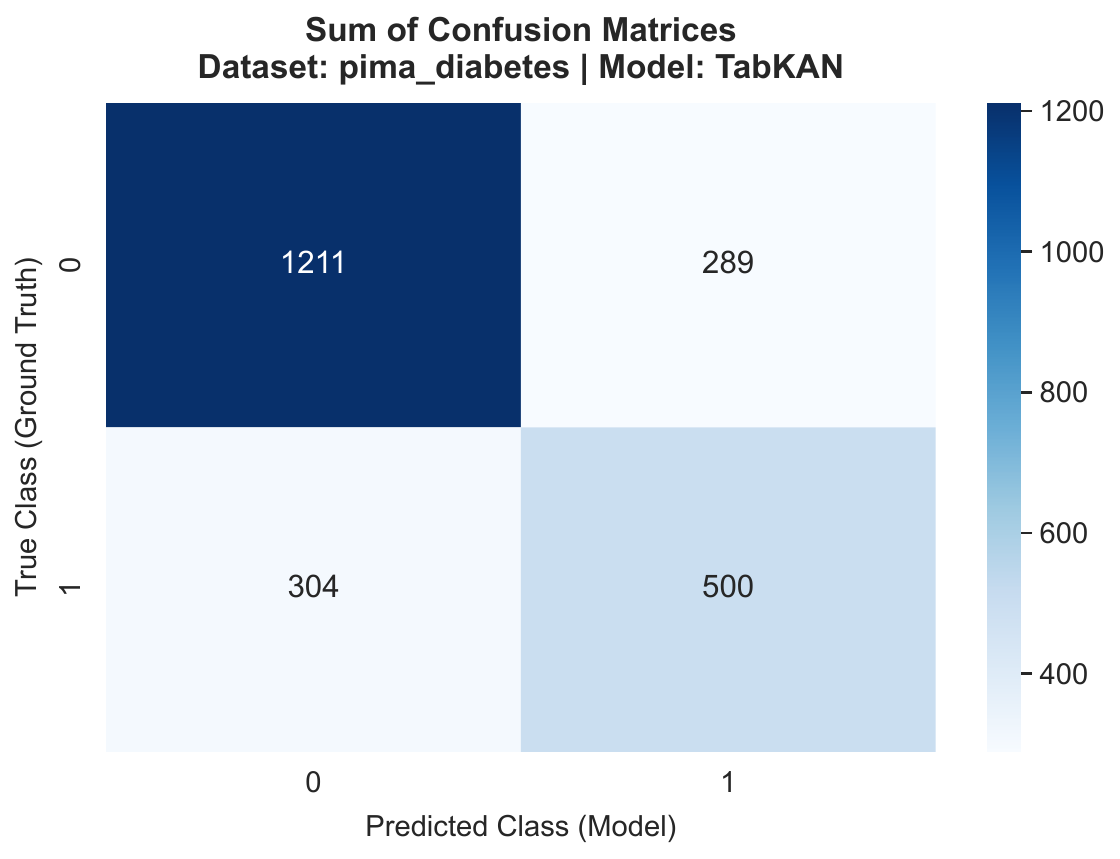}
    \end{subfigure}
    
    \vspace{3mm}
    % Row 4: Breast Cancer
    \begin{subfigure}[b]{0.48\textwidth}
        \centering
        \includegraphics[width=0.8\textwidth]{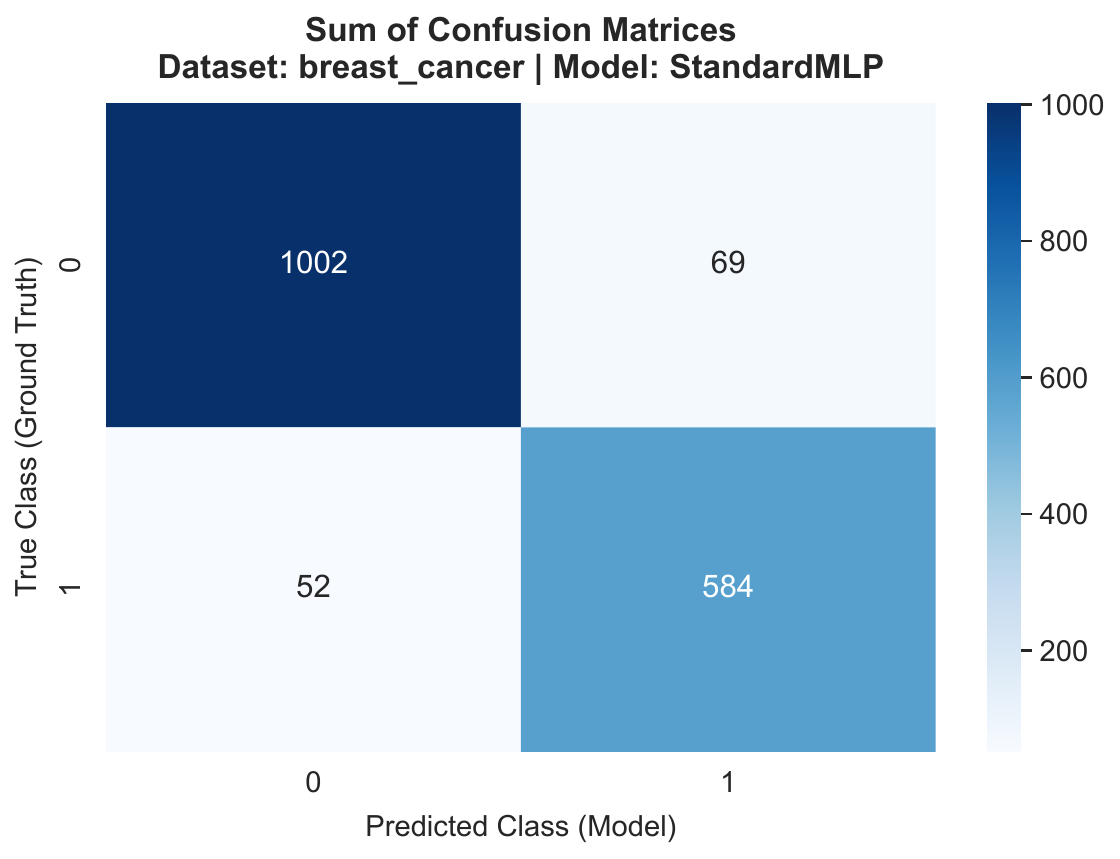}
    \end{subfigure}
    \hfill
    \begin{subfigure}[b]{0.48\textwidth}
        \centering
        \includegraphics[width=0.8\textwidth]{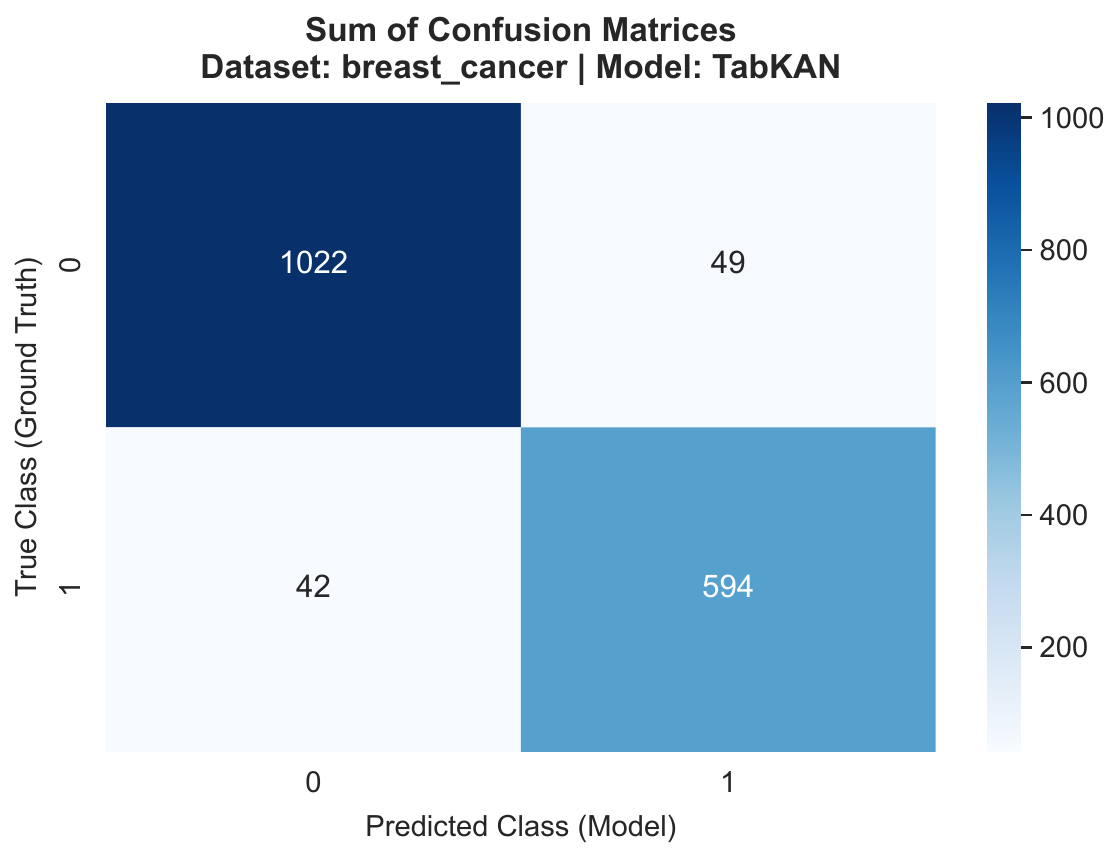}
    \end{subfigure}
    \caption{Confusion Matrix Gallery comparing \textsc{StandardMLP} against representative KAN variants (separate columns). Rows from top to bottom: Cervical Cancer ($6.4\%$ positive prevalence), Cardiotocography ($8.3\%$ minority class), Pima Diabetes, and Breast Cancer.}
    \label{fig:app_cm_grid}
\end{figure*}

To inspect class-specific prediction distributions, Fig.~\ref{fig:app_cm_grid} compares confusion matrices between \textsc{StandardMLP} and the top-performing KAN variant on four benchmark datasets.

On cohorts with high class imbalance, such as Cervical Cancer and Cardiotocography, \textsc{TabKAN} and \textsc{WavKAN} demonstrate improved retrieval of minority positive instances. In Cervical Cancer, \textsc{TabKAN} achieves higher precision on positive biopsy outcomes while reducing false alarms. Similarly, in Cardiotocography, \textsc{WavKAN} achieves higher pathological class recall ($96.9\%$ vs.\ $92.7\%$). In balanced settings, both architectures exhibit nearly identical error distributions, indicating consistent classification behavior across diverse pathologies.

% ====================================================================
% TABLES AND FIGURES
% ====================================================================

\clearpage
\begin{table*}[p]
    \centering
    
    \caption{Summary of Predictive Performance (Grand Mean $\pm$ Std across all datasets and folds)}
    \label{tab:grand_summary_auprc}
    \vspace{1ex}
    \renewcommand{\arraystretch}{1.25}
    
    \resizebox{\textwidth}{!}{%
    \begin{tabular}{lcccccc}
        \toprule
        \textbf{Architecture} & \textbf{AUPRC} & \textbf{AUROC} & \textbf{MCC} & \textbf{F1-Score} & \textbf{Balanced Accuracy} & \textbf{Brier Score} \\
        \midrule
        TabKAN       & $0.9061 \pm 0.1262$ & $0.9481 \pm 0.0651$ & $0.7890 \pm 0.1959$ & $0.8597 \pm 0.1565$ & $0.8858 \pm 0.1062$ & $0.0580 \pm 0.0547$ \\
        StandardMLP  & $0.9028 \pm 0.1278$ & $0.9544 \pm 0.0611$ & $0.7869 \pm 0.1995$ & $0.8543 \pm 0.1549$ & $0.8873 \pm 0.1038$ & $0.0585 \pm 0.0542$ \\
        FastKAN      & $0.9013 \pm 0.1323$ & $0.9480 \pm 0.0630$ & $0.7819 \pm 0.1950$ & $0.8624 \pm 0.1393$ & $0.8844 \pm 0.1008$ & $0.0600 \pm 0.0559$ \\
        ReLUKAN      & $0.8999 \pm 0.1358$ & $0.9464 \pm 0.0642$ & $0.7762 \pm 0.2126$ & $0.8581 \pm 0.1470$ & $0.8833 \pm 0.1093$ & $0.0606 \pm 0.0558$ \\
        TaylorKAN    & $0.8992 \pm 0.1313$ & $0.9470 \pm 0.0634$ & $0.7869 \pm 0.2005$ & $0.8637 \pm 0.1340$ & $0.8917 \pm 0.1000$ & $0.0606 \pm 0.0560$ \\
        WavKAN       & $0.8980 \pm 0.1384$ & $0.9471 \pm 0.0646$ & $0.7858 \pm 0.1985$ & $0.8597 \pm 0.1520$ & $0.8867 \pm 0.1026$ & $0.0591 \pm 0.0569$ \\
        GramKAN      & $0.8962 \pm 0.1392$ & $0.9451 \pm 0.0670$ & $0.7866 \pm 0.2129$ & $0.8681 \pm 0.1309$ & $0.8945 \pm 0.1065$ & $0.0615 \pm 0.0589$ \\
        JacobiKAN    & $0.8952 \pm 0.1384$ & $0.9434 \pm 0.0679$ & $0.7888 \pm 0.1993$ & $0.8688 \pm 0.1323$ & $0.8940 \pm 0.1000$ & $0.0624 \pm 0.0583$ \\
        RandomForest & $0.8923 \pm 0.1455$ & $0.9484 \pm 0.0610$ & $0.7621 \pm 0.1981$ & $0.8458 \pm 0.1578$ & $0.8703 \pm 0.1051$ & $0.0665 \pm 0.0523$ \\
        ChebyKAN     & $0.8911 \pm 0.1465$ & $0.9394 \pm 0.0789$ & $0.7658 \pm 0.2421$ & $0.8442 \pm 0.1779$ & $0.8804 \pm 0.1277$ & $0.0634 \pm 0.0628$ \\
        LegendreKAN  & $0.8894 \pm 0.1466$ & $0.9420 \pm 0.0730$ & $0.7838 \pm 0.2285$ & $0.8559 \pm 0.1710$ & $0.8919 \pm 0.1180$ & $0.0611 \pm 0.0628$ \\
        \bottomrule
    \end{tabular}%
    }

    \vspace{1.5ex}
    \raggedright \footnotesize \textit{Note:} Acronyms and evaluation metrics: AUPRC = Area Under the Precision-Recall Curve; AUROC = Area Under the Receiver Operating Characteristic Curve; MCC = Matthews Correlation Coefficient; F1-Score = Macro-averaged harmonic mean of precision and recall; Balanced Accuracy = Macro-averaged arithmetic mean of sensitivity and specificity; Brier Score = Quadratic scoring rule measuring mean squared difference between predicted probabilities and binary outcomes (lower values indicate superior probability calibration). Values denote grand mean $\pm$ standard deviation aggregated across $N=105$ cross-validation evaluations ($15$~independent folds across $7$ benchmark datasets). Bold typography highlights the optimal mean score per metric column.
    
    \vspace{4em}
    
    \caption{Computational Complexity and Latency Profiling (Grand Mean $\pm$ Std across all datasets and folds)}
    \label{tab:computational_complexity}
    \vspace{1ex}
    
    \resizebox{\textwidth}{!}{
    \begin{tabular}{lcccc}
        \toprule
        \textbf{Architecture} & \textbf{Trainable Params} & \textbf{Inference (ms/sample)} & \textbf{Latency Ratio vs.\ MLP} & \textbf{Train Time (s/fold)} \\
        \midrule
        StandardMLP  & $8{,}273 \pm 5{,}210$    & $0.0611 \pm 0.0165$ & $1.00\times$ & $1.2240 \pm 1.1226$ \\
        ReLUKAN      & $73{,}906 \pm 35{,}963$  & $0.0735 \pm 0.0256$ & $1.20\times$ & $2.0434 \pm 1.6109$ \\
        FastKAN      & $39{,}143 \pm 17{,}119$  & $0.0745 \pm 0.0244$ & $1.22\times$ & $1.6566 \pm 0.9358$ \\
        TabKAN       & $56{,}990 \pm 26{,}956$  & $0.0752 \pm 0.0260$ & $1.23\times$ & $2.0367 \pm 1.5131$ \\
        TaylorKAN    & $23{,}885 \pm 8{,}101$   & $0.0816 \pm 0.0254$ & $1.34\times$ & $1.8621 \pm 1.5215$ \\
        WavKAN       & $103{,}347 \pm 52{,}500$ & $0.0826 \pm 0.0273$ & $1.35\times$ & $2.2309 \pm 2.3698$ \\
        ChebyKAN     & $23{,}361 \pm 7{,}550$   & $0.0881 \pm 0.0297$ & $1.44\times$ & $1.6416 \pm 1.3843$ \\
        GramKAN      & $23{,}678 \pm 7{,}909$   & $0.0882 \pm 0.0285$ & $1.44\times$ & $1.8618 \pm 1.5921$ \\
        LegendreKAN  & $23{,}622 \pm 7{,}876$   & $0.0885 \pm 0.0295$ & $1.45\times$ & $1.7302 \pm 1.3761$ \\
        JacobiKAN    & $22{,}892 \pm 8{,}454$   & $0.0906 \pm 0.0298$ & $1.48\times$ & $1.6060 \pm 1.2573$ \\
        RandomForest & ---                      & $0.1178 \pm 0.0658$ & $1.93\times$ & $0.0746 \pm 0.0521$ \\
        \bottomrule
    \end{tabular}
    }

    \vspace{1.5ex}
    \raggedright \footnotesize \textit{Note:} Architectural complexity and runtime metrics: Trainable Params = Mean total count of learnable network parameters ($|\theta|$); Inference = Mean per-sample inference latency in milliseconds (ms), measured with batch size 1 on Apple Silicon MPS hardware; Latency Ratio = Relative latency factor normalized to the StandardMLP baseline ($1.00\times$); Train Time = Mean wall-clock training duration per cross-validation fold in seconds (s). Values denote grand means $\pm$ standard deviation across all $N=105$ cross-validation folds.
\end{table*}

\clearpage
\thispagestyle{empty}

\clearpage

\clearpage

\clearpage

\end{document}